\documentclass[lettersize,journal]{IEEEtran}
\usepackage{amsmath,amsfonts}
\usepackage{algorithmic}
\usepackage{algorithm}
\usepackage{array}
\usepackage[caption=false,font=normalsize,labelfont=sf,textfont=sf]{subfig}
\usepackage{textcomp}
\usepackage{stfloats}
\usepackage{url}
\usepackage{verbatim}
\usepackage{graphicx}
\usepackage{cite}
\usepackage{bm}
\usepackage{array}
\usepackage{tabularx}
\usepackage{algorithm}
\usepackage{multirow}
\usepackage{booktabs}
\usepackage{array} 
\usepackage{bm}
\usepackage{amssymb}
\usepackage{bbding}
\usepackage{pifont}
\usepackage{wasysym}
\usepackage{amssymb}
\usepackage{graphicx}
\usepackage{makecell}
\usepackage{color}
\usepackage{caption}
\usepackage{xcolor}
\usepackage{subcaption}
\usepackage{multicol}
\usepackage{hyperref}
\hypersetup{hidelinks,
	colorlinks=true,
	allcolors=black,
	pdfstartview=Fit,
	breaklinks=true}
\begin{document}
\title{A Cognitive-Based Trajectory Prediction Approach for Autonomous Driving}

\author{Haicheng Liao$^{*}$, Yongkang Li$^{*}$, Zhenning Li$^{\dag}$, Chengyue Wang, Zhiyong Cui, Shengbo Eben Li,  \IEEEmembership{Senior Member, ~IEEE}, and Chengzhong Xu$^{\dag}$, \IEEEmembership{Fellow, ~IEEE}
\thanks{*\,Authors contributed equally; \dag\,Corresponding author.} \thanks{Haicheng Liao, Yongkang Li, Zhenning Li, and Chengzhong Xu are with the State Key Laboratory of Internet of Things for Smart City and the Department of Computer and Information Science, University of Macau, Macau. Zhiyong Cui is with the School of Transportation Science and Engineering, Beihang University, Beijing. Shengbo Eben Li is with the School of Vehicle and Mobility, Tsinghua University, Beijing, China. Zhenning Li is also with the Department of Civil and Environmental Engineering, University of Macau, Macau. E-mails: {zhenningli, czxu}@um.edu.mo 
}
\thanks{This research is supported by the Science and Technology Development Fund of Macau SAR (File no. 0021/2022/ITP, 0081/2022/A2), and University of Macau (SRG2023-00037-IOTSC).}
}



\maketitle

\begin{abstract}
In autonomous vehicle (AV) technology, the ability to accurately predict the movements of surrounding vehicles is paramount for ensuring safety and operational efficiency. Incorporating human decision-making insights enables AVs to more effectively anticipate the potential actions of other vehicles, significantly improving prediction accuracy and responsiveness in dynamic environments. This paper introduces the Human-Like Trajectory Prediction (HLTP) model, which adopts a teacher-student knowledge distillation framework inspired by human cognitive processes. The HLTP model incorporates a sophisticated teacher-student knowledge distillation framework. The ``teacher'' model, equipped with an adaptive visual sector, mimics the visual processing of the human brain, particularly the functions of the occipital and temporal lobes. The ``student'' model focuses on real-time interaction and decision-making, drawing parallels to prefrontal and parietal cortex functions. This approach allows for dynamic adaptation to changing driving scenarios, capturing essential perceptual cues for accurate prediction. Evaluated using the Macao Connected and Autonomous Driving (MoCAD) dataset, along with the NGSIM and HighD benchmarks, HLTP demonstrates superior performance compared to existing models, particularly in challenging environments with incomplete data. The project page is available at \hypersetup{hidelinks}\href{https://github.com/Petrichor625/HLTP}{\color{purple}{Github}}.
\end{abstract}

\begin{IEEEkeywords}
Autonomous Driving, Trajectory Prediction, Cognitive Modeling, Knowledge Distillation, Interaction Understanding
\end{IEEEkeywords}

\section{Introduction}
\IEEEPARstart{A}ccurate trajectory prediction is central to advancing autonomous vehicle (AV) technology, requiring more than just an analysis of vehicular dynamics and environmental factors. It demands a deep understanding of decision-making processes, which traditional deep-learning models have yet to fully capture \cite{huang2022survey}. Despite the leaps made in vehicular pattern recognition and environmental data processing through advanced deep-learning frameworks, a notable gap persists in the emulation of complex, human-like cognitive functions. Current models, although proficient in handling structured data, often falter in scenarios demanding the adaptability and foresight characteristic of human drivers. Bridging this gap calls for an integrative approach that blends the precision of machine learning algorithms with deep insights derived from cognitive and behavioral studies of human drivers. 

\begin{figure}[t]
    \centering
    \includegraphics[width=\linewidth]{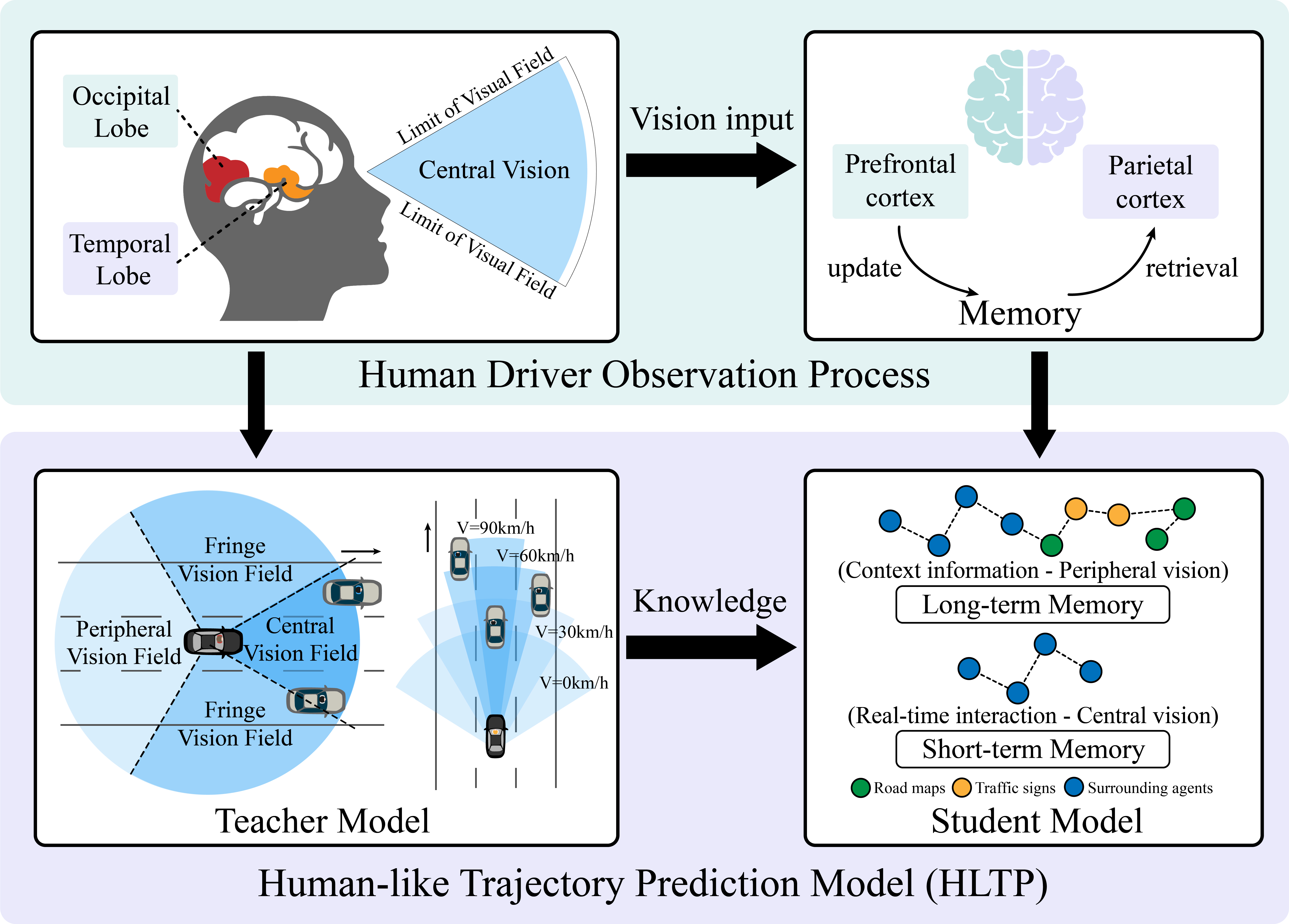}
    \caption{llustration of the HLTP Model. { The ``teacher" model integrates an adaptive visual sector and surround-aware encoder, mirroring the occipital and temporal lobes' roles in visual processing. The ``student" model emphasizes real-time interaction and decision-making, akin to the prefrontal and parietal cortex functions. These components collectively enable HLTP to replicate the intricate visual and cognitive tasks of human drivers, thereby enhancing trajectory prediction.}}
    \label{fig:1}
\end{figure}

As depicted in Figure \ref{fig:1}, human driving expertise stems from a sophisticated network of neural processes. The brain's capacity to integrate sensory information, especially visual cues, and to make instantaneous decisions based on this integration, represents a remarkable cognitive feat. This involves various brain regions, each playing a specific, interconnected role \cite{8818311}. The occipital and temporal lobes, responsible for visual processing, work in concert with the prefrontal and parietal cortex, which are central to decision-making and spatial reasoning \cite{miller2016key}. It is this harmonious interplay that allows drivers to anticipate potential hazards, adjust to sudden changes in the environment, and make informed decisions based on a blend of current sensory input and past experience \cite{gao2020multiple}.

{ 
Drawing inspiration from these cognitive mechanisms, our Human-Like Trajectory Prediction (HLTP) model aims to replicate the essence of human decision-making. The model does not merely seek to forecast trajectories based on historical data but strives to incorporate the dynamism and context-awareness inherent in human cognition. Our study initially focuses on the dynamic allocation of visual attention by human drivers, who focus primarily on the central vision field and pay less attention to the peripheral vision fields, especially during maneuvers such as lane changes, merging, or speed adjustments. The driver's ``vision sector'' adapts dynamically with speed: it narrows at high speed for focused attention ahead, and widens at lower speeds to capture more peripheral information, which is critical for slower driving. Therefore, HLTP introduces an adaptive visual sector that mimics this speed-dependent adaptation with a novel pooling mechanism. This mechanism enhances the model's ability to process and interpret visual information, similar to the human visual system, making it adept at recognizing and responding to important details in different driving environments.

Importantly, we propose a heterogeneous knowledge distillation architecture. The ``teacher" component of HLTP is designed to mimic the brain's visual processing faculties, employing a neural network architecture that dynamically prioritizes and interprets incoming visual data. This aspect of the model is analogous to the way the occipital and temporal lobes filter, process, and relay visual information to other parts of the brain. The ``student" component of HLTP, in turn, represents the decision-making and reasoning aspects of the brain, particularly the functions of the prefrontal and parietal cortex. This part of the model synthesizes the processed visual information, integrating it with spatial awareness and contextual data to make real-time predictive decisions. The goal is to create a model that not only processes information as the human brain does but also adapts and responds to new information with similar flexibility and accuracy. The HLTP model, therefore, is not just an exercise in technical innovation; it is an exploration into the cognitive realms of the human brain, an attempt to harness the complex interplay of neural processes for enhancing AVs' decision-making capabilities. Overall, the contributions of HLTP are multifaceted:

\begin{itemize}
    \item HLTP introduces a novel vision-aware pooling mechanism that incorporates an adaptive visual sector to mimic the dynamics of human visual attention.  Enhanced by a surround-aware encoder and a shift-window attention block, HLTP is able to adeptly capture the critical perceptual cues from diverse scenes.


    \item HTLP employs a cognitive-inspired teacher-student knowledge distillation framework, improving prediction accuracy and robustness, especially in complex scenarios, by efficiently assimilating intricate patterns of visual attention and spatial awareness.
 
    \item The introduction of the Macao Connected and Autonomous Driving (MoCAD) dataset, featuring a right-hand-drive system, offers a novel urban driving context for trajectory prediction studies.
\end{itemize}
}

This paper is organized as follows: In Section \ref{Related Work}, we review the relevant literature in the field.  Section \ref{Method} delves into the intricacies of trajectory prediction and presents our proposed model. Our training methods for this model are detailed in Section \ref{Training_1}. Section \ref{Experiment} evaluates the performance of our model on the NGSIM, HighD, and MoCAD datasets. Finally, in section \ref{Conclusion}, we offer concluding thoughts.


\section{Related Work}\label{Related Work}
\subsection{Trajectory Prediction}
{ 
In the realm of AD, deep learning has catalyzed the adoption of various neural network architectures. Prior studies have explored Recurrent Neural Networks (RNNs) \cite{9852308,quan2021holistic}, Graph Neural Networks (GNNs) \cite{9864070,li2019grip}, attention mechanisms \cite{10179171,chen2022intention}, generative models \cite{wu2023graph,yang2022predicting} and Transformers framework \cite{li2023context,zhou2023query} to enhance spatial and temporal feature extraction from trajectory data. In recent years, some research has focused on solving trajectory prediction tasks from a human cognitive perspective. Notable research, such as BAT \cite{liao2023bat}, applies graph theory to capture human driving behavior. However, it lacks consideration for human driver observation process. Furthermore, Li et al. \cite{li2023context} include visual regions in their model to represent driver perception, but they fail to account for the dynamic adaptation of the vision sector with changing driving speeds and the human memory storage mechanism.
Concurrently, some studies \cite{10268996,kamenev2022predictionnet,chen2023two,liao2024human} focus on improving safety in trajectory predictions, while others \cite{hu2023planningoriented} develop end-to-end models incorporating multimodality. In addition, some recent studies have ventured into the realm of large language models (LLMs) \cite{hu2023planningoriented,liao2023gpt} such as GPT and BLIP, and exploited their ability to output future position distributions. However, the computational intensity of existing models and the lack of human-centered driving cognition pose challenges for real-time application and safety. To address this, we introduce a lightweight yet robust model that mimics human driver observation process and memory storage mechanism, enhancing real-time trajectory prediction capabilities.}
\subsection{Knowledge Distillation}
{ 
Knowledge distillation, originally proposed by Hinton et al. \cite{hinton2015distilling}, involves a complex ``teacher'' model transferring knowledge to a simpler ``student'' model. Initially aimed at model compression, its applications have expanded to initialize pre-trained models \cite{romero2014fitnets}, improve generalization \cite{zagoruyko2016paying}, and increase accuracy \cite{furlanello2018born}, while reducing computational and time requirements \cite{bhardwaj2019efficient}. {   However, its applications in trajectory prediction, especially in the field of autonomous driving, are not yet widespread.} Novel strategies such as Li et al. \cite{9857598} and Monti et al. \cite{monti2022many} allow student models to use fewer observations, while Wsip \cite{wang2023wsip} introduces a wave superposition-inspired social pooling mechanism in this field. Our study integrates human cognitive processes into this paradigm, allowing the teacher model to emulate human reasoning and transfer this knowledge to the student model. This allows for a lightweight model for trajectory prediction that mimics human driving behavior.}

\begin{table*}[htbp]
\centering
\caption{Notation of the main vectors, variables and parameters involved in this paper and their meanings.}
\begin{multicols}{2}
\begin{tabularx}{\columnwidth}{cX}
    \toprule
\textbf{Notation} & \textbf{Meaning} \\
 \midrule
 $\bm{X}$ & 2D position coordinates of the target vehicle and its surrounding vehicles in a traffic scene\\
 $\bm{Y}$ & Predicted future trajectory of target vehicle\\
 $\bm{W}^Q$ & Query weight matrix of the shift-window attention mechanism\\
 $\bm{W}^K$ & Key weight matrix of the shift-window attention mechanism\\
 $\bm{W}^V$ & Value weight matrix of the shift-window attention mechanism\\
 $q_{mn}^i$ & Attention subgraph matrices of the  shift window A\\
 $k_{mn}^i$ & Attention subgraph matrices of the shift window B\\
 $\bm{S}$ & Visual vectors of scene representation\\
  $M$ & Context matrices of scene representation\\
$\bm{S}_{\Delta p}$ &  Relative position of the target vehicle to its neighboring vehicle\\
$\bm{S}_{\Delta s}$ &  Relative velocity of the target vehicle to its neighboring vehicle\\
$\bm{S}_{\Delta a}$ &  Relative acceleration of the target vehicle to its neighboring vehicle\\
$M_{\Delta s}$ & Velocity differences among the surrounding vehicles around adjacent times\\
$M_{\Delta \theta}$ & Direction angle differences among the surrounding vehicles around adjacent times\\
 $\tilde{\bm{S}}$ & Output of the vision-aware pooling mechanism\\
 $\bm{C}$ & Probability of different maneuvers\\

\bottomrule
\end{tabularx}

\begin{tabularx}{\columnwidth}{cX}
\toprule
\textbf{Notation} & \textbf{Meaning} \\
 \midrule
   $H_{\textit{vision}}$ &Visual weight matrix \\ 
    $\bm{O}^i_{\textit{SWA}}$&Output of the $i$th shift-window attention block\\
 $\bm{O}_{\textit{vis}}$  &Output of the shift-window attention mechanism \\
  $\mathbf{O}_{\textit{sur}}$ &Outputs of the surround-aware encoder\\
$\mathbf{O}_{\textit{vis}}$ &Output of the teacher encoder \\
$\mathbf{O}_{\textit{tea}}$ &Output of the ``teacher'' model \\
$\mathbf{O}_{\textit{stu}}$ &Output of the ``student'' model\\
$stride_{x}, stride_{y}$ & Shift strides of the shift window\\
 $\alpha$ & Ratio hyperparameter of knowledge distillation\\
 $T$ & Temperature hyperparameter of knowledge distillation\\
 $\mathcal{L}^N$ & Negative Log-Likelihood loss function\\
 $\mathcal{L}^M$ & Mean Squared Error loss function\\
 $Y_t^{tea}$ & Output of the ``teacher'' model\\
 $Y_t^{stu}$ & Output of the ``student'' model\\
 $\hat{p}^{0}_{t}$ & Ground-truth 2D position coordinates of the target vehicle\\ 
 $p^{0}_{t}$ & 2D position coordinates of the target vehicle predicted by the ``teacher'' model \\
 $\tilde{p}^{0}_{t}$ & 2D position coordinates of the target vehicle predicted by the ``student'' model\\
 $\sigma_1, \sigma_2, \sigma_t, \sigma_d$ & Learnable uncertainty variances for knowledge distillation modulation \\
 \bottomrule
\end{tabularx}
\end{multicols}
\label{table_variables}%
\end{table*}

\section{METHODOLOGY}\label{Method}
\begin{figure*}[t]
  \centering
  \includegraphics[width=0.9\linewidth]{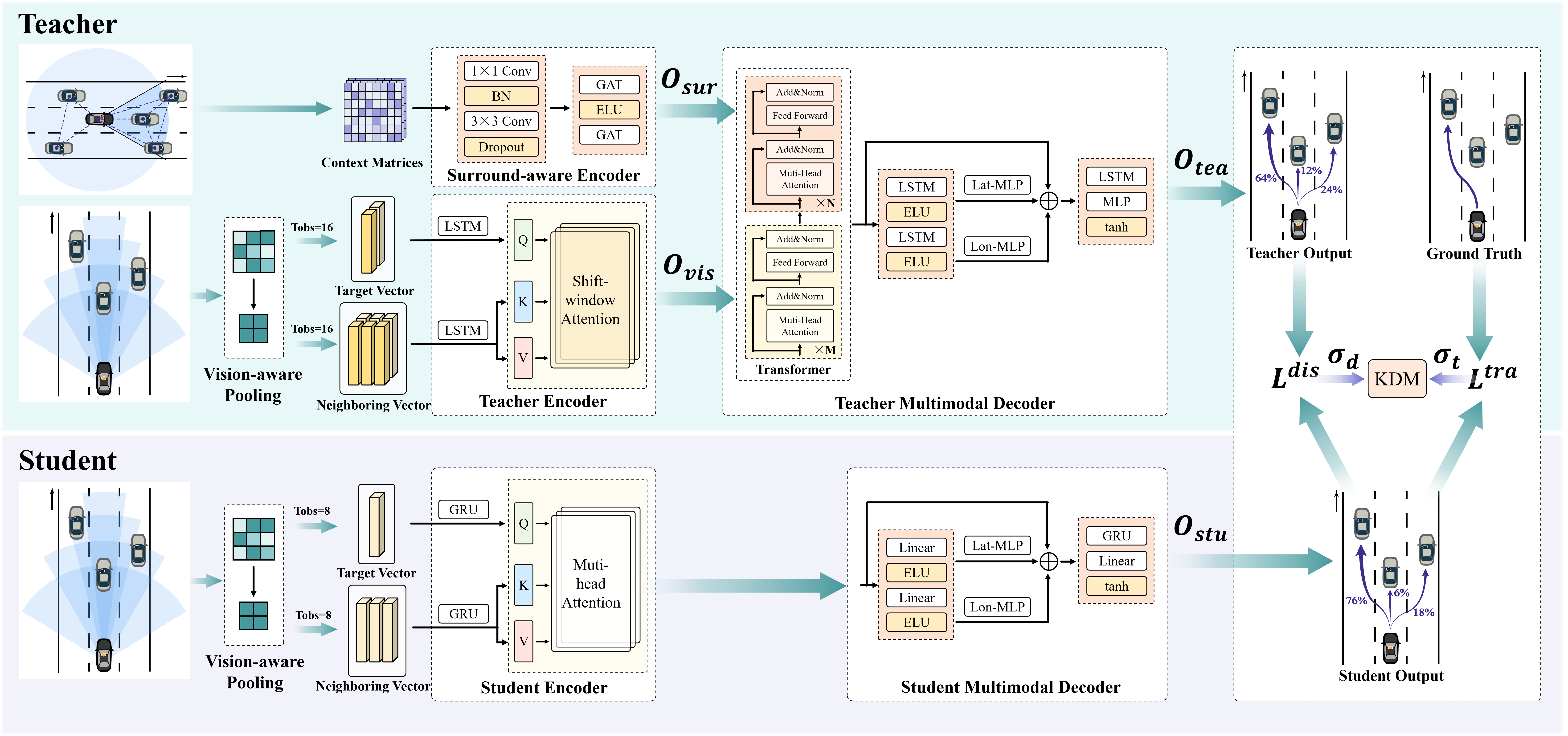}
  \caption{Overall ``teacher-student'' architecture of the HLTP. { The Surround-aware Encoder and the Teacher Encoder within the ``teacher'' model process visual vectors and context matrices to produce surround-aware and visual-aware vectors, respectively. These vectors are then fed into the Teacher Multimodal Decoder, which enables the prediction of different potential maneuvers for the target vehicle, each with associated probabilities. The ``student'' model acquires knowledge from the ``teacher'' model using a Knowledge Distillation Modulation (KDM) training strategy.}}

  \label{fig1}
\end{figure*}
\subsection{Problem Formulation}
The primary aim of this study is to predict the trajectories of all the surrounding vehicles of the AV in a mixed autonomy environment, with each vehicle being referred to as a \textit{target vehicle} during prediction. At any time $t$, the state of the $i$th vehicle can be denoted as $p^{i}_{t}$, where $p^{i}_{t}=(x^{i}_{t}, y^{i}_{t})$ represents the 2D trajectory coordinate. Given the trajectory coordinate data of the target vehicle (superscript 0) and all its surrounding traffic vehicles (superscripts from 1 to $n$) in a traffic scene in the interval $[1, T_{obs}]$, denoted as $\bm{X}=\{p^{0: n}_{t}\}_{t=1}^{T_{obs}} \in \mathbb{R}^{(n+1) \times T_{obs} \times 2}$, the model aims to predict a probabilistic multi-modal distribution over the future trajectory of the target vehicle, expressed as $P(\bm{Y}|\bm{X})$. Here, $\bm{Y}=\{\bm{y}^{0}_{t}\}_{t=T_{obs}+1}^{T_{f}}  \in \mathbb{R}^{T_{f} \times 2}$ is the predictive future trajectory coordinates of the target vehicle over a time horizon $T_{f}$, and $\bm{y}^{0}_{t}=\{(\tilde{p}^{0,1}_{t}; \tilde{c}^{0,1}_{t}),
(\tilde{p}^{0,2}_{t};\tilde{c}^{0,2}_{t}),\cdots, (\tilde{p}^{0,C}_{t};\tilde{c}^{0,C}_{t})\}$ encompasses both the potential trajectory and its associated maneuver likelihood ($\Sigma^{1}_{\bm{C}} {c}^{0,i}_{t}=1$), with $\bm{C}$ denoting the total number of potential trajectories predicted. For the sake of clarity, the main variables and parameters covered in this paper are listed in Table \ref{table_variables}.

\subsection{Model Architecture}
Figure \ref{fig1} illustrates the architecture of HLTP, which features a novel pooling mechanism with an adaptive visual sector. This sector dynamically adapts, similar to human attention allocation, to capture important cues in different traffic situations. Additionally, the model also employs a teacher-student heterogeneous network distillation approach for human-like trajectory prediction.
i) The ``teacher'' model emulates the human visual observation process by integrating a shift-window attention block (SWA) to mimic central visual focus, along with a surround-aware encoder to capture peripheral vision aspects during maneuvers.
ii) The ``student'' model incorporates a human-like vision-aware pooling mechanism and uses a lightweight framework to efficiently represent the spatio-temporal dynamics among traffic participants. Under the guidance of the ``teacher'', the ``student'' strives to bridge the knowledge gap in understanding and emulating human driving behavior, ultimately aiming to improve the accuracy of its predictions, while requiring fewer input observations.

\subsection{Scene Representation}
HLTP describes scenes by focusing on the relative positions of traffic vehicles, aligning with human spatial understanding. It preprocesses historical data into two key spatial forms: \\\textbf{1) Visual vectors $\bm{S}$:} These vectors are constructed to encapsulate the relative position, velocity, and acceleration of the target vehicle to its neighboring vehicles. Represented as $\bm{S}=\{\bm{S}_{\Delta p}, \bm{S}_{\Delta s}, \bm{S}_{\Delta a}\} $, they fall within the dimensional space of $\mathbb{R}^{(n+1) \times T_{obs} \times 4}$; \textbf{2) Context matrices $M$:} These matrices provide a detailed description of speed and direction angle differences among the surrounding vehicles. Defined as $M=\{M_{\Delta s}, M_{\Delta \theta}\} \in \mathbb{R}^{(n+1) \times T_{obs} \times 2}$. They provide a comprehensive view of the relative motion dynamics within the traffic scene. 

These representations $\bm{S}$ and $M$, capturing the geometric nuances of the traffic environment, are then fed into the vision-aware pooling mechanism and teacher-student knowledge distillation framework of HLTP for further feature extraction.

\subsection{Vision-aware Pooling Mechanism}
Research \cite{tucker2021speeding} indicates that human drivers, limited by their brain's working memory, effectively process information from only a few external vehicles, prioritizing the central visual field, particularly in high-risk frontal collision scenarios. Furthermore,  a driver's visual sector, defined by its radius and angle, dynamically adjusts with speed: narrowing at higher speeds for focused attention ahead and widening at lower speeds for broader peripheral awareness.

Inspired by this, HLTP introduces a vision-aware pooling mechanism that emulates the adaptive visual attention of human drivers. It features an innovative adaptive visual sector that adjusts the field of view based on vehicle speed, enabling intuitive and context-sensitive trajectory prediction. This approach departs from models that assume a uniform distribution of attention. Instead, we introduce a visual weight matrix 
$H$ that adapts to represent the driver's shifting focus across the central visual field at different speeds. We establish speed thresholds at $0 km/h$, $30 km/h$, $60 km/h$, and $90 km/h$, each corresponding to different visual sectors. { 
Specifically, different speed thresholds correspond to different central angles of the field of view, as shown in Figure \ref{fig:1}. We assign different learnable weights, denoted $w_c$ and $w_n$, to vehicles inside and outside the central field of view, with initial values set to $w_c=1$ and $w_n=0.2$. Vehicles within the central field of view are assigned higher weights to indicate their greater importance, which is adaptively adjusted in real-time according to different traffic scenarios.}
These thresholds allow different weights to be applied to the model based on the speed of the vehicle, resulting in a more refined understanding of attention allocation during driving. The integration of the visual weight matrix $H_{\textit{vision}}$ with the input visual vectors $\bm{S}$ by element-wise multiplication. Formally,
\begin{equation}
\tilde{\bm{S}} = H_{\textit{vision}} \odot \bm{S},
\end{equation}
This pooling mechanism culminates in the generation of vision-aware vectors $\tilde{\bm{S}} \in \mathbb{R}^{(n+1) \times T_{obs} \times d_s}$ that encapsulate the varying focus and attention patterns of human drivers.

\subsection{Teacher Model}
The ``teacher"  model integrates Teacher Encoder and Surround-aware Encoder to closely emulate human visual perception, mirroring the retinal processing of the human drivers. As an enhancement, it further employs a transformer framework in the decoder to effectively extract spatio-temporal interactions and contextual features. This hierarchical learning process equips the model with advanced knowledge, which is subsequently used to guide the ``student'' model to make human-like trajectory predictions with improved accuracy.

\subsubsection{Teacher Encoder}
The encoder processes central field-of-vision information into vision-aware vectors $\bm{\tilde{S}}$ that are critical for real-time road conditions and immediate obstacles, significantly influencing driving decisions and reactions. In real-world driving, the human brain, with its limited processing capacity, prioritizes information to facilitate efficient decision-making. As a result, the trajectory of a human-driven vehicle is mostly determined by recent traffic scenarios, as the human brain vividly remembers short-term events that influence driving behaviors. In addition, the receptive fields (RFs) of the human visual cortex are limited and typically focus on the most immediate traffic scenes based on neural stimulus intensity. This further implies that human drivers are more inclined to focus on and be mostly influenced by the movements of recent traffic scenarios, as these are perceived as immediate and relevant to their driving decisions.

Furthermore, we introduce a novel shift-window attention block inspired by \cite{liu2021swin} that involves extracting vectors for the target vehicle $\bm{F}_{tar}$ and its neighbors $\bm{F}_{nbr}$ from $\bm{\tilde{S}}$ and embedding them via the LSTM-ELU-Linear structure for high-level vector production.

\begin{figure}
    \centering
    \includegraphics[width=\linewidth]{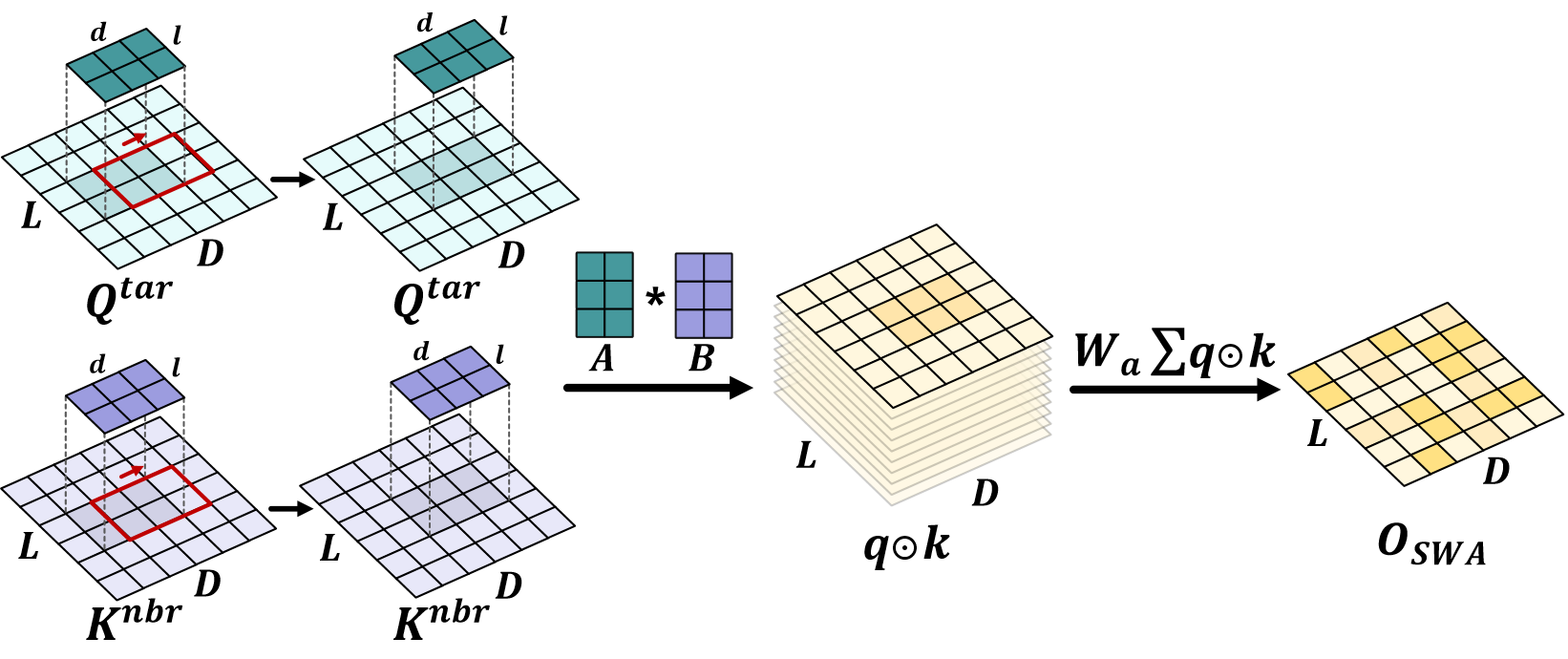}
    \caption{Visualization of the shift-window function in SWA. A window shifting technique is used to capture features by moving windows across key and query vectors, similar to convolution. Overlapping shifted windows, highlighted in red, connect with prior layers, promoting interaction and information sharing between query and key vectors.}
    \label{fig:mdcam2}
\end{figure}
{ Specifically, SWA is built on the multi-head attention mechanism that computes the attention weight within the local windows, while the skip connections are applied for training stability and efficiency. It projects the embedded target vehicle $\bm{\tilde{F}}_{tar}$, neighboring vehicles $\bm{\tilde{F}}_{nbr}$, and the vision-aware $\bm{\tilde{S}}$ vectors into the query \(\mathbf{Q}\), key \(\mathbf{K}\), value \(\mathbf{V}\) representations using Multilayer Perceptrons (MLP) with different embedding dimensions $d_{q}$,$d_{k}$,$d_{v}$, respectively:
\begin{equation}
\mathbf{Q}=\mathbf{W}^{Q} \bm{\tilde{F}}_{tar}, \quad \mathbf{K}=\mathbf{W}^{K} \bm{\tilde{F}}_{nbr}, \quad \mathbf{V}=\mathbf{W}^{V} \bm{\tilde{S}},
\end{equation}
where $\mathbf{W}^{Q}, \mathbf{W}^{K}, \mathbf{W}^{V}$ are the learnable weight matrices. }Then, $\mathbf{Q} \in \mathbb{R}^{(n+1) \times d_{q}}, \mathbf{K} \in \mathbb{R}^{(n+1) \times d_{k}}$ and $\mathbf{V} \in \mathbb{R}^{(n+1) \times d_{v}}$ are divided into $H$ multiple heads, denoted $\mathbf{Q}^{i} \in \mathbb{R}^{(n+1) \times \frac{d_{q}}{H}}$, $\mathbf{K}^{i} \in \mathbb{R}^{(n+1) \times \frac{d_{k}}{H}}$, and $\mathbf{V}^{i} \in \mathbb{R}^{(n+1) \times \frac{d_{v}}{H}}, i \in [1,H]$. The $i$th attention head $\textit{head}^{i}$ and the output $\bm{O}_{\textit{vis}}$ of the shift-window attention mechanism can be defined mathematically as:
\vspace{-5pt}
{\footnotesize
\begin{flalign}
{\bm{O}_{\textit{vis}}}  = \sum_{i=1}^{H} {\textit{head}^{i}} =\sum_{i=1}^{H}  { \phi_{\textit {softmax}}}\left(\frac{\phi_{\textit {shift}}(\mathbf{Q}^{i}, \mathbf{K}^{i})}{\sqrt{d_k}}\right) \odot \tanh(\mathbf{V}^{i}),
\end{flalign}}
where $\phi_{\textit {shift}}$ represents the shift-window function.
Figure \ref{fig:mdcam2} illustrates our approach of treating local windows as the human visual system's finite receptive fields, akin to convolutional kernels. We employ a shift-window method to compute attention weights within these windows, linking them and using their slight overlap for efficient feature aggregation and updating. For $\mathbf{Q}^i$ and $\mathbf{K}^i$ with dimension $L \times D$, we establish unique shift windows $A$ and $B$, each with dimension $l \times d$, considerably smaller than $\mathbf{Q}^i$ and $\mathbf{K}^i$. Both windows have fixed shift steps $stride_{x}$ and $stride_{y}$ for the dimension $x$ and $y$, respectively. Local attention weights are calculated within these windows, which are then sequentially shifted based on their strides. We define $q^i_{mn}$ and $k^i_{mn}$ as attention subgraph matrices for windows A and B, with $m$ and $n$ indicating their respective displacements in the $x$ and $y$ directions. This process produces the output $\bm{O}^i_{\textit{SWA}}$:
\begin{equation}
\begin{aligned}
\bm{O}^i_{\textit{SWA}}=\phi_{\textit {shift}}(\mathbf{Q}^i, \mathbf{K}^i) = \mathbf{W}_a \sum_{m=0}^{z_{x}} \sum_{n=0}^{z_{y}} q^{i}_{mn}\odot k^{i}_{mn}, 
\end{aligned}
\end{equation}
where the weighted product of these maps is computed using the weight matrix $\mathbf{W}_a$. To ensure synchronized traversal of these windows, it is essential that $z_x=z_y=\frac{L-l}{stride_x}+1 = \frac{D-d}{stride_y}+1$. This SWA allows for precise modeling and processing of attention, closely mimicking the selective and dynamic focusing of human vision in complex driving environments.\\

\subsubsection{Surround-aware Encoder}
Human drivers focus on their central vision while continuously monitoring their peripheral vision through the side and rear-view mirrors to understand their surroundings, including nearby vehicles, pedestrians, and road conditions. To replicate this peripheral monitoring, especially during maneuvers, we introduce the surround-aware encoder. It processes a quarter of the time-segmented matrices $M  \in \mathbb{R}^{(n+1) \times \frac{T_{obs}}{4} \times 2}$ with a $1\times 1$ convolutional layer for channel expansion, followed by a $3\times 3$ layer for specific feature extraction, incorporating batch normalization and dropout for robustness. Enhanced with Graph Attention Networks (GAT) and ELU activation, it produces surround-aware vectors $\mathbf{O}_{\textit{sur}}$. \\

\subsubsection{Teacher Multimodal Decoder} 
The combined outputs of the surround-aware $\mathbf{O}_{\textit{sur}}$ and the teacher $\mathbf{O}_{\textit{vis}}$ encoders are fed into a transformer for advanced spatio-temporal interaction analysis, generating hidden states $\mathbf{I}$. The decoder of the ``teacher" model, based on a Gaussian Mixture Model (GMM), accounts for uncertainty in trajectory prediction by evaluating multiple possible maneuvers and their probabilities.

To provide a formal description, the future trajectories can be predicted hierarchically in a Bayesian framework on two hierarchical levels: (1) At each time step, the probability of different maneuvers $\bm{C}$ of the target vehicle is determined; (2) Subsequently, the detailed trajectories of the vehicle conditional on each maneuver are generated within a predefined distributional form. In accordance with the characteristics of the driver's actions during driving, the possible maneuvers of the vehicles are decomposed into a combination of two distinctive sub-maneuvers, comprising the position-wise sub-maneuver ${c}_{p}$ and the velocity-wise sub-maneuver ${c}_{v}$. Additionally, the position-wise sub-maneuver includes three discrete driver choices regarding position changes, namely left lane change, right lane change, and lane keeping. Meanwhile, the speed-wise sub-maneuver includes three distinct decisions, including accelerating, braking, and maintaining speed, which serve as branch categories.

Conditional on the estimated maneuvers $\bm{C}$ in the first layer, the probability distribution of the multimodal trajectory predictions is assumed to follow a Gaussian distribution:
\begin{equation}
    P_{\bm{\Omega}} (\bm{Y}|\bm{C},\bm{X}) = N(\bm{Y}|\mu(\bm{X}),\Sigma(\bm{X})),
\end{equation}
where $\bm{X}$ denotes the input of the proposed model. Moreover, $\bm{\Omega}=\left[\Omega^{t+1}, \ldots,\Omega^{t+t_{f}}\right]$ are the estimable parameters of the distribution, and $\Omega^{t}=[\mu^{t},\Sigma^{t}]$ are the mean and variance of the distribution of predicted trajectory point at $t$. Correspondingly, in the second layer, the multimodal predictions are formulated as a Gaussian Mixture Model, i.e.,
\begin{equation}\label{eq.6}
    P(\bm{Y}|\bm{X})=\sum_{\forall i} P\left(c_{i}|\bm{X}\right) P_{\bm{\Omega}}\left(\bm{Y}|c_{i},\bm{X}\right),
\end{equation}
where $c_{i}$ is the $i$-th element in $\bm{C}$.

This multimodal structure not only provides different predictions, but also quantifies their confidence levels, which supports decision-making amidst unpredictability in prediction. The hidden states $\mathbf{I}$ are then processed through softmax activation and an MLP, forming a probability distribution $\mathbf{O}_{\textit{tea}}$ over potential future trajectories. This approach balances precision and adaptability, which is critical for dynamic and uncertain driving environments.

\subsection{Student Model}
{ The ``student'' model mimics human memory storage strategies by prioritizing real-time interactions and central visual field data, similar to a driver's focus during navigation. Information from peripheral vision, such as road signs and maps, provides broader context but receives less attention, reflecting the division between short-term and long-term memory in human cognition.} This design ensures that the ``student'' model closely mirrors the behavior of the ``teacher'' model, balancing immediate detail with overarching context in decision-making. It focuses on short-term input, processing a limited number of short-term, vision-aware vectors $\tilde{\bm{S}}$ ($T_{obs} =8$), using a lightweight architectural design for efficient learning. In a departure from the more complex Transformer framework of the ``teacher'' model, the ``student'' model employs the GRU for its decoder to produce a multimodal prediction distribution $\mathbf{O}_{stu}$.  By distilling knowledge from the ``teacher'' model, the ``student'' model can make effective predictions even when constrained by limited observational data.

\section{Training}\label{Training_1}
\subsection{Teacher Training} 
For the ``teacher'' model, we follow a standard protocol, using 3 seconds of observed trajectory for input ($T_\textit{obs}=16$) and predicting a 5-second future trajectory ($T_\textit{f}=25$) with a sampling rate of 2. To extract complex knowledge from the dataset, we allow slight over-fitting during training. To address the potentially detrimental effects of misclassified maneuver types on trajectory prediction accuracy and robustness, we adopt a multitask learning approach for teacher loss, which can be decoupled as Mean Squared Error (MSE) loss $\mathcal{L}^{M}$ and Negative Log-Likelihood (NLL) loss  $\mathcal{L}^{N}$. 

{ The NLL Loss function is commonly used in probabilistic models. For a specific data point, given the model's predicted probability distribution $P(\bm{Y}|\bm{X})$, where $\bm{Y}$ represents the output future trajectory and $\bm{X}$ denotes the input features, the NLL Loss $\mathcal{L}^N$ is defined: $\mathcal{L}^N=-log(P(\bm{Y}|\bm{X}))$.

In our trajectory prediction process, we treat the output trajectory as a bivariate Gaussian distribution consisting of the horizontal coordinate $ x $ and the vertical coordinate $ y $. Given the ground-truth 2D position coordinate $ \hat{{p}}^{gt}_{t}= (\hat{x}_t^{gt}, \hat{y}_t^{gt}) $, and the final predicted trajectory set by the teacher model is $ p_t^{tea} = (x_t^{tea}, y_t^{tea}) $, the possibility $P(\hat{p}_t^{gt}|p_t^{tea})$ is defined as: 
\begin{equation}\label{nll}
\footnotesize
\begin{aligned}
P(\hat{p}_t^{gt}|p_t^{tea}) = \frac{1}{2\pi \sigma_x \sigma_y \sqrt{1-\rho^2}}
exp\left\{-\frac{1}{2(1-\rho^2)}\left[\left(\frac{\hat{x}_t^{gt}-\mu_x}{\sigma_x}\right)^2\right. \right. \\
-2\rho \frac{\hat{x}_t^{gt}-\mu_x}{\sigma_x}\frac{\hat{y}_t^{gt}-\mu_y}{\sigma_y}
\left.\left.+\left(\frac{\hat{y}_t^{gt}-\mu_y}{\sigma_y}\right)^2 \right]\right \},
\end{aligned}
\end{equation}
where $P(\hat{p}_t^{gt}|p_t^{tea}) =(\tilde{c}^{0,1}_{t}, \tilde{c}^{0,2}_{t}, \tilde{c}^{0,3}_{t}, \cdots, \tilde{c}^{0,C}_{t})$, $\mu_x$, $\mu_y$, $\sigma_x$, $\sigma_y$ are the means and variances of $\hat{x}^{gt}_t$, $\hat{y}^{gt}_t$, $\rho$ is the correlation coefficient between $\hat{x}_t^{gt}$ and $\hat{y}_t^{gt}$.}

Moreover, we calculate the teacher loss using the trajectory prediction set of ``teacher'' model $ \bm{Y}_t^{tea}=\{({p}^{0,1}_{t}; {c}^{0,1}_{t}),({p}^{0,2}_{t};{c}^{0,2}_{t}),\cdots, ({p}^{0,C}_{t};{c}^{0,C}_{t})\}$ and the ground-truth set $ \bm{\hat{Y}}_t^{gt}= \{(\hat{{p}}^{0}_{t}; \hat{{c}}^{0}_{t})\}$ with their corresponding maneuver probability distributions, $t \in [T_{obs}+1, T_f]$. This loss function, denoted as $\mathcal{L}^{tea}$, is derived by a one-to-one calculation between the predicted and actual trajectories:
\begin{equation}
\footnotesize
\begin{array}{r}
\mathcal{L}^{tea}=\sum_{t}^{T_\textit{f}}\sum_{i}^{C}\left[\mathcal{L}^{M}\left({p}_{t}^{0,i}, \hat{{p}}^{0}_{t}\right)+\mathcal{L}^{N}\left(\bm{Y}_t^{tea}, \bm{\hat{Y}}_t^{gt}\right)\right]
\end{array}
\end{equation}

\subsection{Student Training}
The ``student'' model is trained to predict 5-second trajectories with fewer input observations. To improve its predictive performance with limited observations, we decouple the student loss $\mathcal{L}^{stu}$ as \textit{track} loss $\mathcal{L}^{tra}$ and \textit{distillation} loss $\mathcal{L}^{dis}$. The track loss, similar to that of the ``teacher'' model, quantifies the discrepancy between the model's predicted trajectories and their ground truths. Mathematically,
\begin{equation}
\footnotesize
\begin{array}{r}
\mathcal{L}^{tra}=\sum_{t}^{T_\textit{f}}\sum_{i}^{C}\left[\mathcal{L}^{M}\left(\tilde{p}_{t}^{0,i}, \hat{{p}}^{0}_{t}\right)+\mathcal{L}^{N}\left(\bm{Y}_t^{stu}, \bm{\hat{Y}}_t^{gt}\right)\right],\\ \\
\textit{such that 
 }\bm{Y}_t^{stu}=\{(\tilde{p}^{0,1}_{t}; \tilde{c}^{0,1}_{t}),
(\tilde{p}^{0,2}_{t};\tilde{c}^{0,2}_{t}),\cdots, (\tilde{p}^{0,C}_{t};\tilde{c}^{0,C}_{t})\},
\end{array}
\end{equation}
Here, $ \bm{Y}_t^{stu}$ represents the output of ``student'' model.
In contrast, the MSE loss used in \textit{distillation} focuses on guiding ``student'' model to mimic the behavior of its ``teacher":
\begin{equation}
\footnotesize
\begin{aligned}
\mathcal{L}^{dis}
&=2\alpha T^2 \sum_{t}^{T_\textit{f}}\sum_{i}^{C}\left[\mathcal{L}^{N}\left(\tilde{P}_{t}^{0,i}, {{P}}^{0,i}_{t}\right)+\mathcal{L}^{N}\left(\tilde{c}^{0,i}_{t}, {c}^{0,i}_{t}\right)\right],
\end{aligned}
\end{equation}
Here, $T$ is the distillation temperature, used to modulate the smoothness of the output probability distribution, and $\alpha$ is a hyperparameter to measure the proportion between the two loss functions. Correspondingly, $\tilde{P}_{t}^{0,i}=\phi_{\textit {softmax}} (\frac{\tilde{p}_{t}^{0,i}}{T})$ denotes the output obtained post-softmax application. 
Eventually, the student loss $\mathcal{L}^{stu}$ is the summation of \textit{track} loss $\mathcal{L}^{tra}$ and  \textit{distillation} loss $\mathcal{L}^{dis} =\mathcal{L}^{tra}+ \mathcal{L}^{dis}$.
\subsection{Knowledge Distillation Modulation}

Given that $L^{stu}$ is composed of sub-loss functions from multiple tasks, determining the proportionality relationships between them poses a challenging problem. 

Both $\mathcal{L}^{tra}$ and $\mathcal{L}^{dis}$ are further decomposed into maneuver loss function $\mathcal{L}_{man}$ and trajectory coordinate loss function $\mathcal{L}_{coor}$:
\begin{equation}
\begin{aligned}
\mathcal{L}^{tra}  = \mathcal{L}^{tra}_{man} + \mathcal{L}^{tra}_{coor},\,\,
\mathcal{L}^{dis}  = \mathcal{L}^{dis}_{man} + \mathcal{L}^{dis}_{coor},
\end{aligned}
\end{equation}
where each part is formulated as follows:
\begin{equation}
\footnotesize
\begin{aligned}
\mathcal{L}^{tra}_{coor}&=\sum_{t}^{T_\textit{f}}\sum_{i}^{C}\mathcal{L}^{M}\left(\tilde{p}_{t}^{0,i}, \hat{{p}}^{0}_{t}\right),\\
\mathcal{L}^{tra}_{man}&=\sum_{t}^{T_\textit{f}}\sum_{i}^{C}\mathcal{L}^{N}\left(\bm{Y}_t^{stu}, \bm{\hat{Y}}_t^{gt}\right),\\
\mathcal{L}^{dis}_{coor}&=2\alpha T^2 \sum_{t}^{T_\textit{f}}\sum_{i}^{C}\mathcal{L}^{N}\left(\tilde{P}_{t}^{0,i}, {{P}}^{0,i}_{t}\right),\\
\mathcal{L}^{dis}_{man}&=2\alpha T^2 \sum_{t}^{T_\textit{f}}\sum_{i}^{C}\mathcal{L}^{N}\left(\tilde{c}^{0,i}_{t}, {c}^{0,i}_{t}\right),\\
\end{aligned}
\end{equation}
Notably, inspired by previous work \cite{kendall2018multitask}, we incorporate a Knowledge Distillation Modulation to weight the \textit{track} loss and the \textit{distillation} loss, with homoscedastic uncertainty. To the best of our knowledge, we are the first to propose a multi-level, multi-task hyperparameter tuning approach to autonomously adjust knowledge distillation hyperparameters during training in this field. A multi-level task is defined where the overarching training loss function is composed of an ensemble of sub-loss functions. Each of these sub-loss functions is further composed of additional sub-loss functions, all of which have some degree of similarity to each other.
\begin{table*}[htbp]
  \centering
  \caption{Comparison of MoCAD with other AV datasets. The datasets highlighted in \textbf{bold} are those that we evaluated specifically for this study. Within the vehicle column, ``Veh.'' represents a range of vehicles, including cars, trucks, trailers, vans, and buses. Meanwhile, ``Ped.'' and ``Bic.'' are shorthand for pedestrians and bicyclists respectively. In the context of Data Type, ``Traj.'' refers to the trajectory coordinates, while ``HD maps'' denotes the high-definition (HD) maps. Furthermore, ``PT'' and ``SG'' in the Location column stand for Pittsburgh and Singapore, correspondingly.}
  \resizebox{\linewidth}{!}{
    \begin{tabular}{ccccccccc}
    \toprule
    Dataset & Year  & vehicle & Scene & Data Type & Tracking Quantity & Sensor & Sampling Rate & Location \\
    \hline
    \textbf{NGSIM} \cite{deo2018convolutional}  & 2006  & Veh. & Highway & Traj. & 1.5 hours & Cemara & 10 Hz & USA \\
    KITTI \cite{geiger2013vision}& 2013  & Bic., Ped. & Urban, Highway & Image, Point cloud & 50 Sequences & Lidar, Camera & 10 Hz & Karlsruhe \\
    \textbf{HighD} \cite{highDdataset}& 2018  & Veh. & Highway & Traj., Lane & 447 hours, 110500 vehicles & Drone & 25 Hz & German \\
    Apolloscape \cite{huang2018apolloscape}& 2018  & Veh., Bic., Ped. & Urban & Traj. & 1000 Km Trajectories & Lidar, Camera & 10 Hz & China \\
    Argovrse \cite{chang2019argoverse}& 2019  & Vehicles & Urban & Traj., HD maps & 320h  & Lidar, Camera & 30 Hz & Miami, PT \\
    RounD \cite{rounDdataset}& 2020  & Veh., Bic., Ped. & Roundabouts & Traj., Lanes & 13746 Tracks & Drone & 25 Hz & German \\
    nuScenes \cite{caesar2020nuscenes}& 2020  & Veh., Bic., Ped. & Urban & Traj., HD maps & 5.5 hours & Lidar, Camera & 12 Hz & Boston, SG \\
    \textbf{MoCAD (Ours)} & 2023  & Veh., Bic., Ped. & Urban, Campus & Traj., HD maps, Point cloud, BEV & 300 hours & Lidar, Camera, Drone  & 10 Hz & Macao\\
   \toprule
    \end{tabular}%
    }
  \label{Datasets}%
\end{table*}%
{ Theoretically, there are two options for multi-level, multi-task hyperparameter tuning in our context:
Option (1): The first level involves multi-task tuning of $\mathcal{L}^{tra}$ and $\mathcal{L}^{dis}$, while the second level involves multi-task tuning of $\mathcal{L}_{man}$ and $\mathcal{L}_{coor}$.
Option (2): The first level deals with multi-task tuning of $\mathcal{L}_{man}$ and $\mathcal{L}_{coor}$, while the second level focuses on $\mathcal{L}^{tra}$ and $\mathcal{L}^{dis}$. We first adopt option (1) using the multi-task tuning approach outlined in \cite{kendall2018multitask} for the first level of the loss function, we derive the following equation:
\begin{equation}
\begin{aligned}
\mathcal{L}^{stu}_1=\frac{1}{2 \sigma_{t}^2} \mathcal{L}^{tra}(\mathbf{W})+\frac{1}{2 \sigma_{d}^2} \mathcal{L}^{dis}(\mathbf{W})+ log\sigma_{t}\sigma_{d},
\end{aligned}
\end{equation}

Then, employing multi-task tuning separately for the second level of the loss functions results in the following formulation:
\begin{equation}
\begin{aligned}
\mathcal{L}^{tra}=\frac{1}{2 \sigma_{1}^2} \mathcal{L}^{tra}_{man}(\mathbf{W})+\frac{1}{2 \sigma_{2}^2} \mathcal{L}^{tra}_{coor}(\mathbf{W})+ log\sigma_{1}\sigma_{2},\\
\mathcal{L}^{dis}=\frac{1}{2 \sigma_{1}^2} \mathcal{L}^{dis}_{man}(\mathbf{W})+\frac{1}{2 \sigma_{2}^2} \mathcal{L}^{dis}_{coor}(\mathbf{W})+ log\sigma_{1}\sigma_{2},\\
\end{aligned}
\end{equation}
where $\sigma_1, \sigma_2, \sigma_t, \sigma_d$ are the learnable uncertainty variances.

Considering the analogous properties of $\mathcal{L}^{tra}_{man}$ with $\mathcal{L}^{dis}_{man}$, $\mathcal{L}^{tra}_{coor}$ with $\mathcal{L}^{dis}_{coor}$, we apply the same variance adjustments, separately. By substituting the above expressions, we derive the following formulation:
\begin{equation}\label{option1}
\footnotesize
\begin{aligned}
\mathcal{L}^{stu}_1(W,\sigma_1,\sigma_2,\sigma_t,\sigma_d)
=\frac{1}{2\sigma^2_t}(\frac{1}{2\sigma_1^2}\mathcal{L}^{tra}_{man}+\frac{1}{2\sigma_2^2}\mathcal{L}^{tra}_{coor})\\
+\frac{1}{2\sigma^2_d}(\frac{1}{2\sigma_1^2}\mathcal{L}^{dis}_{man}+\frac{1}{2\sigma_2^2}\mathcal{L}^{dis}_{coor})
+log\sigma_1\sigma_2(\frac{1}{2\sigma^2_t}+\frac{1}{2\sigma^2_d})
+log\sigma_t\sigma_d,\\
\end{aligned}
\end{equation}
Similarly, by adopting option (2), we obtain the following:
\begin{equation}\label{option2}
\footnotesize
\begin{aligned}
\mathcal{L}^{stu}_2(W,\sigma_1,\sigma_2,\sigma_t,\sigma_d)
=\frac{1}{2\sigma_1^2}(\frac{1}{2\sigma^2_t}\mathcal{L}^{tra}_{man}+\frac{1}{2\sigma^2_d}\mathcal{L}^{dis}_{coor})\\
+\frac{1}{2\sigma_2^2}(\frac{1}{2\sigma^2_t}\mathcal{L}^{dis}_{man}+\frac{1}{2\sigma^2_d}\mathcal{L}^{dis}_{coor})
+log\sigma_t\sigma_d(\frac{1}{2\sigma_1^2}+\frac{1}{2\sigma_2^2})
+log\sigma_1\sigma_2,
\end{aligned}
\end{equation}

The difference between Eq. \ref{option1} and Eq. \ref{option2} merely lies in their respective constant terms. Specifically, the expression $log\sigma_1\sigma_2(\frac{1}{2\sigma^2_t}+\frac{1}{2\sigma^2_d})
+log\sigma_t\sigma_d$ is not equal to to $log\sigma_t\sigma_d(\frac{1}{2\sigma_1^2}+\frac{1}{2\sigma_2^2})
+log\sigma_1\sigma_2$. This demonstrates the non-equivalence of the two multi-task tuning approaches. Therefore, we optimize the formula by omitting the constant term and replacing it with $\log \sigma_1 + \log \sigma_2 + \log \sigma_t + \log \sigma_d$, or alternatively as $\log \sigma_1\sigma_2\sigma_t\sigma_d$. This approach not only introduces regularization by the uncertainty variances $\sigma_1, \sigma_2, \sigma_t, \sigma_d$ and makes different options share the same expression, but also simplifies the formula, making the training process easier.}
\begin{figure}[htbp]
  \centering
\includegraphics[width=0.7\linewidth]{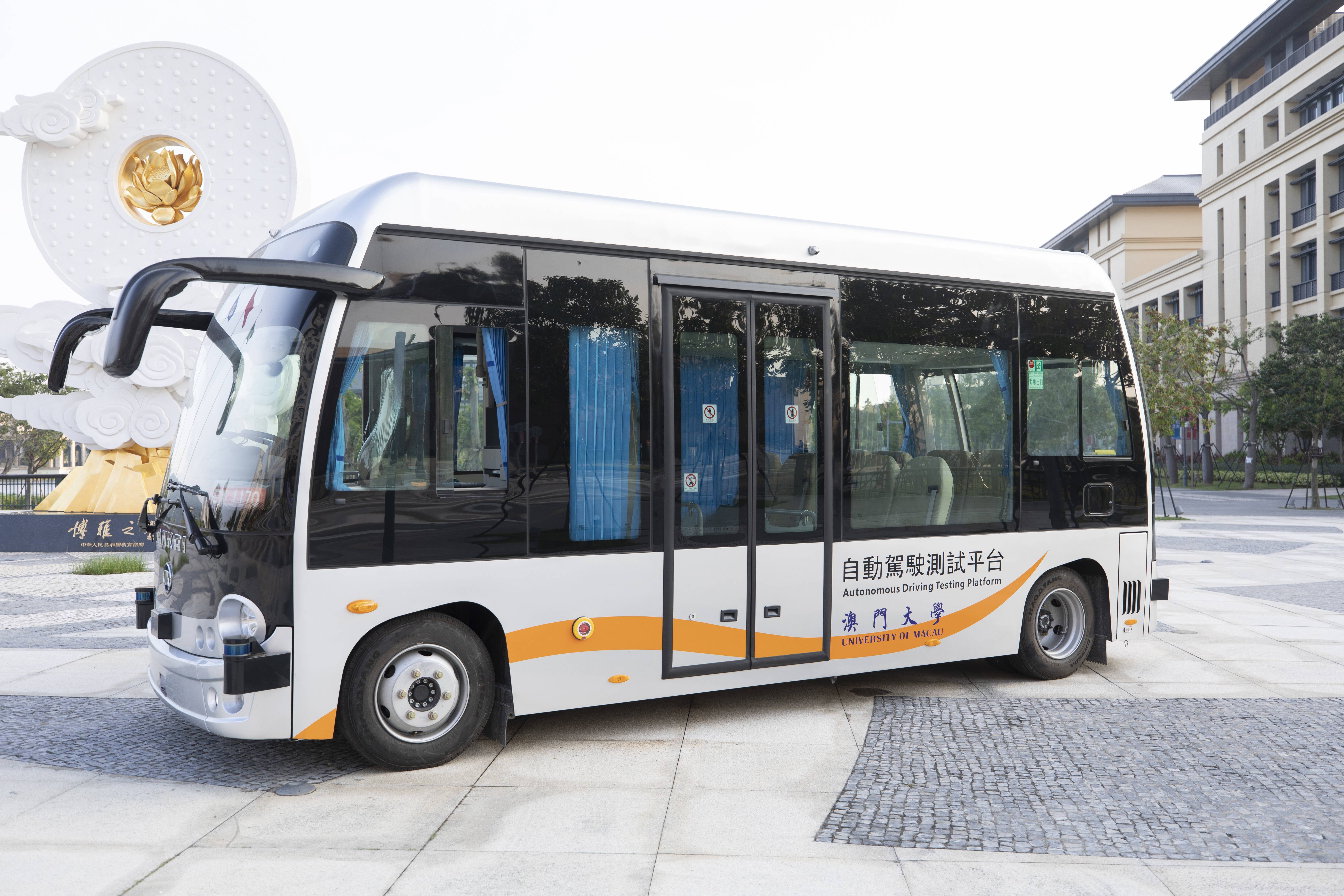} 
  \caption{Autonomous driving testing platform of University of Macau.}
  \label{bus} 
\end{figure}
\begin{table*}[htbp]
  \centering
     \caption{Evaluation results for our proposed model and the other SOTA baselines without using HD maps in the NGSIM, HighD and MoCAD datasets over a different horizon. Note: RMSE (m) is the evaluation metric, where lower values indicate better performance, with some not specifying (``-''). \textbf{Bold} and \underline{underlined} values represent the best and second-best performance in each category. ``AVG'' indicates the average value of the RMSE.}\label{Table1}
     \setlength{\tabcolsep}{7mm}
   \resizebox{0.85\linewidth}{!}{
    \begin{tabular}{c|ccccccc}
    \toprule
    \multicolumn{1}{c}{\multirow{2}[2]{*}{Dataset}} & \multirow{2}[3]{*}{Model} & \multicolumn{6}{c}{Prediction Horizon (s)} \\
\cmidrule{3-8}    \multicolumn{1}{c}{} &       & 1     & 2     & 3     & 4     & 5 & AVG\\
      \midrule
    \multirow{12}[22]{*}{NGSIM}
          & S-GAN \cite{gupta2018social} & 0.57  & 1.32  & 2.22  & 3.26  & 4.40 & 2.35 \\
          & CS-LSTM \cite{deo2018convolutional} & 0.61  & 1.27  & 2.09  & 3.10  & 4.37 & 2.29 \\
          & MATF-GAN \cite{zhao2019multi} & 0.66  & 1.34  & 2.08  & 2.97  & 4.13 & 2.22 \\
          & IMM-KF \cite{lefkopoulos2020interaction} & 0.58  & 1.36  & 2.28  & 3.37  & 4.55 & 2.43 \\
          & MFP \cite{tang2019multiple} & 0.54  & 1.16  & 1.89  & 2.75  & 3.78 & 2.02 \\
          & DRBP\cite{gao2023dual} & 1.18  & 2.83  & 4.22  & 5.82  & -  & 3.51 \\
          & WSiP \cite{wang2023wsip} & 0.56  & 1.23  & 2.05  & 3.08  & 4.34 & 2.25 \\
          & CF-LSTM \cite{xie2021congestion} & 0.55  & 1.10  & 1.78  & 2.73  & 3.82  & 1.99 \\
          & MHA-LSTM \cite{messaoud2021attention} & 0.41  & 1.01  & 1.74  & 2.67  & 3.83 & 1.91 \\
          & HMNet \cite{xue2021hierarchical} & 0.50  & 1.13  & 1.89  & 2.85  & 4.04 & 2.08 \\
          & TS-GAN \cite{wang2022multi} & 0.60  & 1.24  & 1.95  & 2.78  & 3.72 & 2.06 \\
          & STDAN \cite{chen2022intention} & \textbf{0.39}  & 0.96  & 1.61  & 2.56 & 3.67 & 1.84 \\ 
            & iNATran (M) \cite{chen2022vehicle} & 0.41  & 1.00  & 1.70  & 2.57  & 3.66  & 1.87  \\ 
          & iNATran  \cite{chen2022vehicle} & \textbf{0.39}  & \underline{0.96}  & 1.61  & 2.42  & 3.43  & 1.76  \\  
          
          & DACR-AMTP \cite{cong2023dacr}& 0.57  & 1.07  & 1.68  & 2.53  & 3.40  & 1.85  \\ 
          & FHIF \cite{zuo2023trajectory} &\underline{0.40}  & 0.98  & 1.66  & 2.52  & 3.63  & 1.84  \\  
               &HLTP                                    & {0.41}    & \textbf{0.91}     & \textbf{1.45} & \textbf{2.09} & \textbf{2.93} & \textbf{1.56}\\ 
    &HLTP (s)                                & 0.48             & {0.99}  & \underline{1.54} & \underline{2.22} & \underline{3.09} &\underline{1.66}\\
      \midrule
    \multirow{7}[20]{*}{HighD}
    &S-GAN \cite{gupta2018social}& 0.30  & 0.78  & 1.46  & 2.34  & 3.41  &1.69 \\
    &WSiP \cite{wang2023wsip}& 0.20  & 0.60  & 1.21  & 2.07  & 3.14 &1.44 \\
    &CS-LSTM(M) \cite{deo2018convolutional}& 0.23  & 0.65  & 1.29  & 2.18  & 3.37  &1.54 \\
    &CS-LSTM \cite{deo2018convolutional}& 0.22  & 0.61  & 1.24  & 2.10  & 3.27 &1.48 \\
    &MHA-LSTM \cite{messaoud2021attention}& 0.19  & 0.55  & 1.10  & 1.84  & 2.78 &1.29 \\
    &MHA-LSTM(+f) \cite{messaoud2021attention}& 0.06  & 0.09  & 0.24  & 0.59  & 1.18 &0.43\\
    &DRBP\cite{gao2023dual}& 0.41  & 0.79  & 1.11  & 1.40  & -  & 0.92\\
    &EA-Net \cite{cai2021environment} & 0.15  & 0.26  & 0.43  & 0.78  & 1.32  &0.59 \\
    &CF-LSTM \cite{xie2021congestion}& 0.18  & 0.42  & 1.07  & 1.72  & 2.44 &1.17  \\
    &STDAN \cite{chen2022intention}& 0.19  & 0.27  & 0.48  & 0.91  & 1.66  &0.70 \\
     &DACR-AMTP \cite{cong2023dacr}& 0.10  & \underline{0.17}  & \underline{0.31}  & {0.54}  & {1.01} &0.42 \\
       &HLTP & \textbf{0.09} & \textbf{0.16} & \textbf{0.29} & \textbf{0.38} & \textbf{0.59} & \textbf{0.30}\\
    &HLTP (s) & \underline{0.11} & 0.23 & 0.32 & \underline{0.42} & \underline{0.68}& \underline{0.35} \\
     \midrule
    \multirow{7}[10]{*}{MoCAD}
    &S-GAN \cite{gupta2018social} & 1.69  & 2.25  & 3.30  & 3.89  & 4.69 &3.16 \\
    &CS-LSTM(M) \cite{deo2018convolutional}& 1.49  & 2.07  & 3.02  & 3.62  & 4.53 &2.95   \\
    &CS-LSTM \cite{deo2018convolutional} & 1.45  & 1.98  & 2.94  & 3.56  & 4.49 &2.88 \\
    &MHA-LSTM \cite{messaoud2021attention} & 1.25  & 1.48  & 2.57  & 3.22  & 4.20  & 2.54\\
    &MHA-LSTM(+f) \cite{messaoud2021attention} & 1.05  & 1.39  & 2.48  & 3.11  & 4.12 &2.43 \\
    &WSiP \cite{wang2023wsip} & 0.70  & 0.87  & 1.70  & 2.56  & 3.47 &1.86  \\
    &CF-LSTM \cite{xie2021congestion} & 0.72  & 0.91  & 1.73  & 2.59  & 3.44 &1.87 \\
    &STDAN \cite{chen2022intention} & \underline{0.62}  & \underline{0.85}  & 1.62  & 2.51  & 3.32 &1.78  \\
       &HLTP  & \textbf{0.55} & \textbf{0.76} & \textbf{1.44} & \textbf{2.39} & \textbf{3.21}&\textbf{1.67} \\
    &HLTP (s)                                & 0.64 & 0.86 & \underline{1.55} & \underline{2.48} & \underline{3.27} &\underline{1.76}\\
    \bottomrule
    \end{tabular}%
  \label{table_overall}%
  }
\end{table*}%
\section{Experiment}\label{Experiment}
\subsection{Datasets} 
{ We evaluate the effectiveness of our model using the three datasets: Next Generation Simulation (NGSIM) \cite{deo2018convolutional}, Highway Drone (HighD) \cite{highDdataset}, and our proposed MoCAD. The MoCAD dataset \cite{liao2023bat} was collected from the first Level 5 autonomous bus in Macau (as shown in Figure \ref{bus}) and has undergone extensive testing and data collection since its deployment in 2020. The data collection period spans over 300 hours and covers various scenarios, including a 5-kilometer campus road dataset, a 25-kilometer dataset covering city and urban roads, and complex open traffic environments captured under different weather conditions, time periods, and traffic densities. As shown in Table \ref{Datasets}, we present a comprehensive comparison of our proposed MoCAD dataset with other datasets for AVS. It is noteworthy that MoCAD stands out with its extensive tracking hours and unique Macao origin, featuring a right-hand-drive system and mandatory left-hand driving, providing rare insights into complex driving patterns for trajectory prediction research.}

\subsection{Experimental Setup}
\subsubsection{Dataset Segmentation}  In this study, all three datasets are treated with a consistent segmentation framework, i.e., each trajectory is divided into 8-second segments, with the first 3 seconds as historical data and the following 5 seconds for evaluation. 

\subsubsection{Metric} { Root Mean Square Error (RMSE) and average RMSE are used as our primary evaluation metrics.}

\subsubsection{Implementation Details}\label{Training} 
Our model is developed using PyTorch and trained to converge on an A40 48G GPU. We use the Adam optimizer along with CosineAnnealingWarmRestarts for scheduling, with a training batch size of 128 and learning rates ranging from $10^{-3}$ to $10^{-5}$. Unless specified, all evaluation results are based on the ``student'' model of HLTP. In addition, we lighten the architecture of the ``student'' model by halving the hidden state dimensions of the GRU and reducing the number of attention heads, termed HLTP (s), for subsequent experiments. 


\subsection{Experiment Results}
\subsubsection{Comparison with the State-of-the-art Baseline}
Our comprehensive evaluation demonstrates HLTP's superior performance compared to SOTA baselines, as detailed in Table \ref{table_overall}. HLTP's ``student'' model, trained on only 1.5 seconds ($T_{obs}=8$) of recent trajectory data, surpasses the standard 3-second data reliance of baseline models. It notably achieves gains of 5.2\% for short-term (2s) and 13.8\% for long-term (5s) predictions on the NGSIM dataset. On the HighD dataset, all models, including HLTP, display fewer inaccuracies due to HighD's precise trajectories and larger data size. While short-term predictions are comparable across models, HLTP excels in long-term forecasting with a 41.6\% RMSE improvement for up to 5 seconds. In complex right-hand-drive environments like urban streets and unstructured roads (MoCAD dataset), HLTP's accuracy gains range from 3.3\% to 11.3\%, underscoring its robustness and adaptability in diverse traffic scenarios.

\subsubsection{Comparing Model Performance and Complexity} 
In our benchmark against SOTA baselines, as detailed in Table \ref{tab:parameters}, HLTP and HLTP (s) models demonstrate superior performance across all metrics, while maintaining a minimal parameter count. Despite limited access to many models' source codes, our analysis focuses on open-source options. Remarkably, HLTP (s) achieves high performance with significantly less complexity, reducing parameters by 71.41\% and 55.89\% compared to WSiP and CS-LSTM, respectively. Furthermore, HLTP (s) efficiently outperforms transformer-based STDAN and CF-LSTM, using 82.34\% and 77.79\% fewer parameters, respectively. This highlights the efficiency and adaptability of our lightweight ``teacher-student" knowledge distillation framework, offering practitioners a balance between accuracy and computational resource requirements.\\
\begin{table}[htbp]
  \centering
  \caption{Comparative evaluation of our model with selected baselines. Emphasizing model complexity via parameter count (Param.). \textbf{Bold} and \underline{underlined} values represent the best and second-best performance in each category.}
    \setlength{\tabcolsep}{4mm}
    \resizebox{\linewidth}{!}{
    \begin{tabular}{c|cccc}
    \toprule
    \multicolumn{1}{c}{\multirow{2}[4]{*}{Model}} & \multirow{2}[4]{*}{Param. (K)} & \multicolumn{3}{c}{Average RMSE (m)} \\
\cmidrule{3-5}  \multicolumn{1}{c}{} &   & NGSIM & HighD & MoCAD  \\
    \midrule 
        CS-LSTM      & 194.92 & 2.29  & 1.49  & 2.88   \\
        CF-LSTM      & 387.10 & 1.99  & 1.17  & 1.88   \\
        WSiP         & 300.76 & 2.25  & 1.44  & 1.86   \\
        STDAN        & 486.82 & 1.87  & 0.70  & 1.78   \\
        {HLTP}        & \underline{149.50} & \textbf{1.56}  & \textbf{0.30}  &   \textbf{1.67} \\
        {HLTP (s)}        & \textbf{85.97} & \underline{1.66} & \underline{0.35} & \underline{1.76} \\
    \bottomrule
    \end{tabular}
    }
  \label{tab:parameters}%
\end{table}%
\subsubsection{Qualitative Results}
\begin{figure}[h!]
    \centering
    \includegraphics[width=1\linewidth]{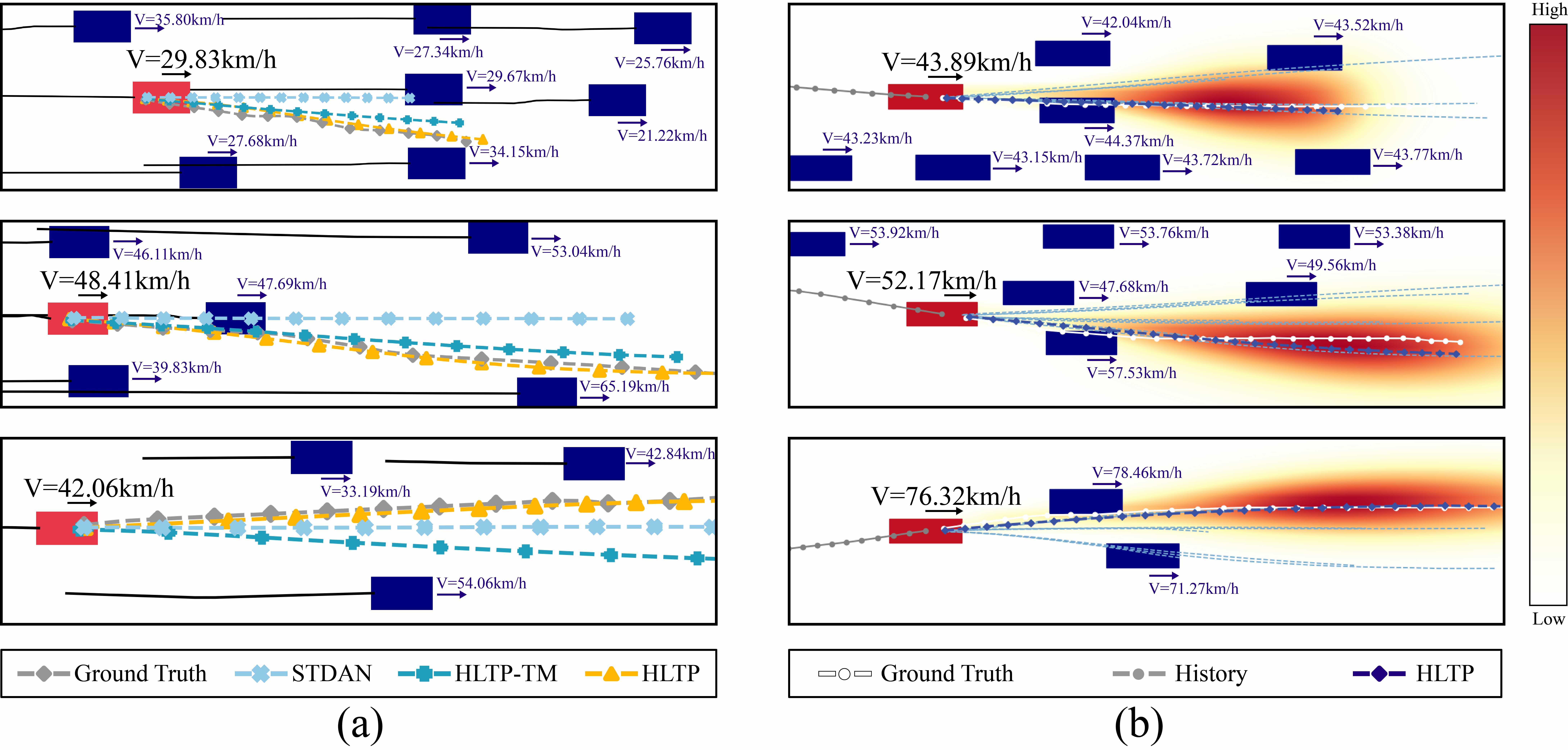}
    \caption{Multimodal probabilistic prediction (a) and visualizations (b) for the target vehicle on NGSIM. Heat maps illustrate the GMM of predictions: brighter areas denote higher probabilities. The target vehicle is marked in {\color{red} red} and its surrounding vehicles in {  blue}.}
    \label{qual}
\end{figure}

{ Figure \ref{qual} showcases the multimodal probabilistic prediction performance of HLTP on the NGSIM dataset, while showing the velocities of the target vehicle and its surrounding vehicles. The heat maps shown represent the Gaussian Mixture Model of predictions in challenging scenes. These visualizations show that the highest probability predictions of our model are very close to the ground truth, indicating its impressive performance.  Additionally, it also visually demonstrates our model's ability to accurately predict complex scenarios such as merging and lane changing, confirming its effectiveness in various traffic situations.  Interestingly, in certain complex scenarios, the trajectory predictions of the ``student'' model exceed the accuracy of the ``teacher'' model. This result illustrates the ability of the ``student'' model to selectively assimilate and refine the knowledge acquired from the ``teacher'' model, effectively ``extracting the essence and discarding the dross''. }

\begin{table*}[!t]
  \centering
  \caption{Ablation studies for core component, different observation input, and missing data.}\label{tab:ablation}%
    \begin{minipage}[t]{0.5\linewidth}
    \centering
    \captionsetup{labelformat=empty,skip=1pt}
        \caption*{{\footnotesize(a) Different methods and components of ablation study.}}\label{(1) }
        \setlength{\tabcolsep}{1.8mm}
        \resizebox{0.9\linewidth}{!}{
            \begin{tabular}{cccccccc}
                \toprule
                \multirow{2}[4]{*}{Components} & \multicolumn{7}{c}{Ablation Methods} \\
                \cmidrule{2-8}          & A     & B     & C     & D     & E  &F &G \\
                \midrule
                Vision-aware Pooling Mechanism & \ding{56} & \ding{52} & \ding{52} & \ding{52} & \ding{52} & \ding{52} & \ding{52} \\
                Surround-aware Encoder & \ding{52} & \ding{56} & \ding{52}& \ding{52}  & \ding{52} & \ding{52}  & \ding{52}\\
                 Transformer  Module& \ding{52} & \ding{52} &  \ding{56} &\ding{52} & \ding{52}  & \ding{52}  & \ding{52}\\
                 Shift-window Attention Block & \ding{52} & \ding{52} & \ding{52} & \ding{56} & \ding{52}  & \ding{52}  & \ding{52}\\
                 Multimodal Decoder & \ding{52} & \ding{52} & \ding{52} & \ding{52}& \ding{56}  & \ding{52}  & \ding{52}\\
                 Knowledge Distillation Modulation & \ding{52} & \ding{52} & \ding{52} & \ding{52}& \ding{52}  & \ding{56}  & \ding{52}\\
                \bottomrule
            \end{tabular}}%
    \end{minipage}%
    \begin{minipage}[t]{0.5\linewidth}
          \centering
          \captionsetup{labelformat=empty,skip=1pt}
          \caption*{{\footnotesize(b) Ablation results for different models. (NGSIM/HighD)}}
          \label{(2)}
           \setlength{\tabcolsep}{1mm}
          \resizebox{0.9\linewidth}{!}{
            \begin{tabular}{cccccccc}
            \toprule
            \multirow{2}[4]{*}{Time (s)} & \multicolumn{6}{c}{Ablation Methods} \\
        \cmidrule{2-7}          & A     & B     & C     & D     & E  &F \\
            \midrule
             1 & 0.39/0.09 & 0.52/0.18 & 0.51/0.10 & 0.40/0.12 & 0.46/0.21 & 0.50/0.10 \\
             2 & 0.93/0.17 & 1.09/0.25 & 1.10/0.15 & 0.95/0.21 & 1.06/0.37 & 1.01/0.19 \\
             3 & 1.57/0.36 & 1.77/0.37 & 1.80/0.28 & 1.58/0.33 & 1.76/0.58 & 1.56/0.34 \\
             4 & 2.34/0.51 & 2.56/0.57 & 2.68/0.46 & 2.31/0.58 & 2.66/0.75 & 2.22/0.47 \\
             5 & 3.28/0.70 & 3.45/0.72 & 3.69/0.69 & 3.23/0.83 & 3.78/1.02 & 3.09/0.68 \\
             AVG & 1.70/0.37 & 1.88/0.42 & 1.96/0.34 & 1.69/0.41 & 1.94/0.57 & 1.68/0.36 \\
            \bottomrule
            \end{tabular}}%
    \end{minipage}%
\end{table*}%
\begin{table*}[!t]
  \footnotesize
  \centering
    \begin{minipage}[t]{0.5\linewidth}
    \centering
    \captionsetup{labelformat=empty,skip=1pt}
    \caption*{{\footnotesize (c) Ablation results for varying observation frames (recent/initial).}}
    \label{(3)}
        \small
        \setlength{\tabcolsep}{1.5mm}
        \resizebox{0.9\linewidth}{!}{
        \begin{tabularx}{\linewidth}{c|cccccc}
        \toprule
        \multicolumn{1}{c}{Time (s)}&$T_{obs}$=4 &$T_{obs}$=6 &$T_{obs}$=10 &$T_{obs}$=12 &$T_{obs}$=16  \\
        \midrule
                1 & 0.42/0.68 & 0.42/0.67 & 0.41/0.56 & 0.42/0.53 & 0.41/0.41 \\ 
                2 & 0.95/1.33 & 0.93/1.28 & 0.92/1.15 & 0.92/1.12 & 0.91/0.91 \\ 
                3 & 1.50/2.04 & 1.47/1.96 & 1.45/1.76 & 1.44/1.69 & 1.43/1.43 \\ 
                4 & 2.17/2.85 & 2.12/2.76 & 2.08/2.50 & 2.07/2.39 & 2.05/2.05 \\ 
                5 & 3.05/3.86 & 2.98/3.75 & 2.91/3.44 & 2.90/3.29 & 2.87/2.87 \\  
        \bottomrule
        \end{tabularx}%
        }
    \end{minipage}%
    \begin{minipage}[t]{0.5\linewidth}
    \centering
    \captionsetup{labelformat=empty,skip=1pt}
    \caption*{{\footnotesize (d) Ablation results for \textit{missing} data.}}
    \label{(4)}
        \small
        \setlength{\tabcolsep}{4.2mm}
        \resizebox{0.9\linewidth}{!}{
        \begin{tabularx}{\linewidth}{c|cccccc}
        \toprule
        \multicolumn{1}{c}{Time (s)}&3-10 &4-11 &5-12 &6-13 &7-14  \\
        \midrule
                1 & 0.43 & 0.44 & 0.45 & 0.45 & 0.48 \\ 
                2 & 0.95 & 0.96 & 0.98 & 0.99 & 1.04 \\ 
                3 & 1.50 & 1.52 & 1.55 & 1.58 & 1.65 \\ 
                4 & 2.16 & 2.19 & 2.24 & 2.29 & 2.37 \\ 
                5 & 3.02 & 3.06 & 3.12 & 3.20 & 3.30 \\
        \bottomrule
        \end{tabularx}%
        }
    \end{minipage}%
\end{table*}%

\subsection{Ablation Studies}
\subsubsection{Ablation Study for Core Components}  Table \ref{tab:ablation}(a) shows that our ablation study evaluates the performance of HLTP using six model variations, each omitting different components. Method A excludes the vision-aware pooling mechanism, using data from all surround vehicles; Method B excludes the surround-aware module; Method C omits the transformer module; Method D replaces SWA with classical attention; Method E omits multimodal probabilistic maneuver prediction; Method F avoids a multitask learning strategy; {  Method G, for comparison, includes all these elements, i.e., the complete proposed model HLTP.}

\begin{table*}[!t]
  \centering
  \caption{Ablation studies for the hyperparameters of the SWA. Top-performing models are \textbf{bolded} in each category.}\label{shift-window}
    \begin{minipage}[t]{0.5\linewidth}
    \centering
    \captionsetup{labelformat=empty,skip=1pt}
        \caption*{{\footnotesize(a) Ablation study on stride of the shift window in SWA.}}\label{(1)b }
        \setlength{\tabcolsep}{0.5mm}
        \resizebox{0.9\linewidth}{!}{
            \begin{tabular}{cccccc}
                \toprule
                \multirow{2}{*}{Time (s)} & $stride_x=8$ & $stride_x=8$ & $stride_x=8$ & $stride_x=8$ & $stride_x=16$ \\
                 & $stride_y=4$  & $stride_y=6$  & $stride_y=8$  & $stride_y=12$ & $stride_y=12$ \\
                \midrule
                1 & 0.43 & \textbf{0.41} & \textbf{0.41} & 0.42 & 0.43 \\ 
                2 & 0.93 & \textbf{0.91} & \textbf{0.91} & 0.92 & 0.94 \\ 
                3 & 1.47 & \textbf{1.43} & \textbf{1.43} & 1.45 & 1.48 \\ 
                4 & 2.09 & \textbf{2.05} & 2.06 & 2.08 & 2.09 \\ 
                5 & 2.92 & \textbf{2.87} & 2.88 & 2.89 & 2.92 \\
                \bottomrule
            \end{tabular}}%
    \end{minipage}%
    \begin{minipage}[t]{0.5\linewidth}
          \centering
          \captionsetup{labelformat=empty,skip=1pt}
          \caption*{{\footnotesize(b) Ablation study on size of the shift window in SWA.}}
          \label{(2)b}
           \setlength{\tabcolsep}{0.5mm}
          \resizebox{0.9\linewidth}{!}{
            \begin{tabular}{cccccc}
            \toprule
            Time (s) & x=32, y=12 & x=32, y=18 & x=32, y=24 & x=32, y=30 & x=32, y=36 \\
            \midrule
                1 & \textbf{0.40} & \textbf{0.40} & 0.41 & 0.42 & 0.42 \\ 
                2 & 0.93 & \textbf{0.91} & \textbf{0.91} & 0.92 & 0.93 \\ 
                3 & 1.46 & 1.44 & \textbf{1.43} & 1.45 & 1.46 \\ 
                4 & 2.10 & 2.08 & \textbf{2.05} & 2.07 & 2.07 \\ 
                5 & 2.94 & 2.91 & \textbf{2.87} & 2.89 & 2.90 \\ 
            \bottomrule
            \end{tabular}}%
    \end{minipage}%
\end{table*}%

The results, presented in Table \ref{tab:ablation}(b), show that all removal methods underperform compared to Method G. The inclusion of the transformer module and a multi-task learning approach in the model significantly improves performance, illustrating their critical role in capturing high-level interactions and evaluating different sources of knowledge for accurate trajectory prediction.  Furthermore, Method E omits the multimodal probabilistic maneuver prediction, which also results in a decrease in performance.  In addition, the inclusion of the vision-aware pooling mechanism significantly improved performance by adaptively adjusting the attention weights for the surrounding vehicles. By incorporating the human driver's observation process, HLTP predicts the target vehicle's trajectory more insightfully. Similarly, employing the Surround-aware Encoder in Method E outperformed Method B without this encoder, highlighting the relevance of contextual information for traffic scenario relationships.  Furthermore, removing the Shift-window Attention block in Method D results in considerable RMSE increases, confirming its vital role in HLTP.

\subsubsection{Ablation Study for Different Input and Distribution}
To explore the impact of varying observation lengths on the HLTP student model, we experimented with different  $T_{obs}=4,6,10,12,16$, as shown in Table \ref{tab:ablation}(c). Furthermore, with $T_{obs}=6$, HLTP exceeds all baselines in prediction accuracy. As the amount of observation data increases ($T_{obs}>8$), we observe a slight reduction in prediction error for HLTP; however, the overall performance improvement is marginal. These findings align with research on human attention and decision-making, which suggests that humans tend to prioritize information that is relevant to their current goals and that is within their immediate visual field \cite{kahneman1973attention}. 
To elucidate the effect of different distributions of historical time points on the model's performance, we experimented with different input time distributions. For instance, in the case of $T_{obs}=4$ we compared the impact of using data from the first to the fourth frames with that of the most recent time points, data from the $(T_{obs}-3)$th to $(T_{obs})$th frames. The results clearly indicate that using the initial moments versus the most recent data significantly affects the model's performance, further validating the premise that recent historical data has the most significant influence on future trajectory predictions.

\subsubsection{Ablation Study for Missing Data}
We introduced a \textit{missing} test set on the NGSIM dataset, focusing on scenarios where half of the historical data is missing. This set is divided into five subsets based on different distributions of missing data. For example, the $3-10$ subset excludes data from frames 3 through 10. We used linear interpolation to fill these gaps. The results in Table \ref{tab:ablation}(d) show that HLTP outperforms all baselines even with 50\% missing data, highlighting its adaptability and deep understanding of traffic dynamics. Moreover, we observed that the closer the missing data is to the prediction window ($6-13$, $7-14$ subsets), the more significant the impact on model performance, reinforcing the importance of recent historical data.

\subsubsection{Ablation Study for the SWA}
We investigate the effect of the size and stride of the shifting window in SWA on the predictive performance of the HLTP. 
As shown in Table \ref{shift-window} (a) and (b), in the proposed SWA block, unintuitively, the best experimental results for ablation studies on window stride and window size are achieved with intermediate parameter values. According to ablation studies, the optimal window size is $x=32, y=24$ with strides of $stride_x=8, stride_y=6$, rather than using a larger sliding window and strides. { We attribute this result to two primary factors: the effect of the number of windows on the complexity of the training, and the effect of the number of windows on the amount of information captured. A smaller $stride_y$ results in a larger number of windows, making it more difficult to train the attention weights $W_a$. Conversely, a larger $stride_y$ reduces the number of windows, thereby limiting the variety of information scales. However, as long as the stride values are within a reasonable range, the union of all windows will still cover the entire dataset, resulting in a small performance penalty. The same principle applies to decreasing the window size. Decreasing the dimension $y$ leads to an increase in the number of windows, thus increasing the training difficulty. Conversely, increasing the dimension $y$ results in fewer windows, thus reducing the amount of information.
Overall, our empirical findings resonate with the observations on human attention and decision-making. This research indicates that people typically prioritize information relevant to their current objectives and within their immediate visual field, as suggested by Kahneman \cite{kahneman1973attention}. In addition, our findings support the idea that an overly broad range of observations can actually impair predictive performance. This may underscore the importance of focusing on the targeted and relevant information data in AD systems.}


\section{Conclusion}\label{Conclusion}
{ In this paper, we propose a novel trajectory prediction model HLTP for AVs that addresses the limitations of previous models in terms of parameter heaviness and limited applicability. 
It is based on a heterogeneous knowledge distillation network, providing a lightweight yet efficient framework that maintains prediction accuracy. Importantly, HLTP adapts to scenarios with missing data and reduced inputs, using layered learning to simulate human observation and make human-like predictions based on fewer input observations. Our empirical results show that HLTP excels in complex traffic scenarios and delivers SOTA performance. Additionally, we present the right-hand-drive system dataset MoCAD as a new benchmark to refine trajectory prediction methodologies in complex traffic scenes.}

\bibliographystyle{IEEEtran}
\bibliography{IEEEabrv,IEEEexample}

\begin{IEEEbiography}
[{\includegraphics[width=1in,height=1.40in, clip,keepaspectratio]{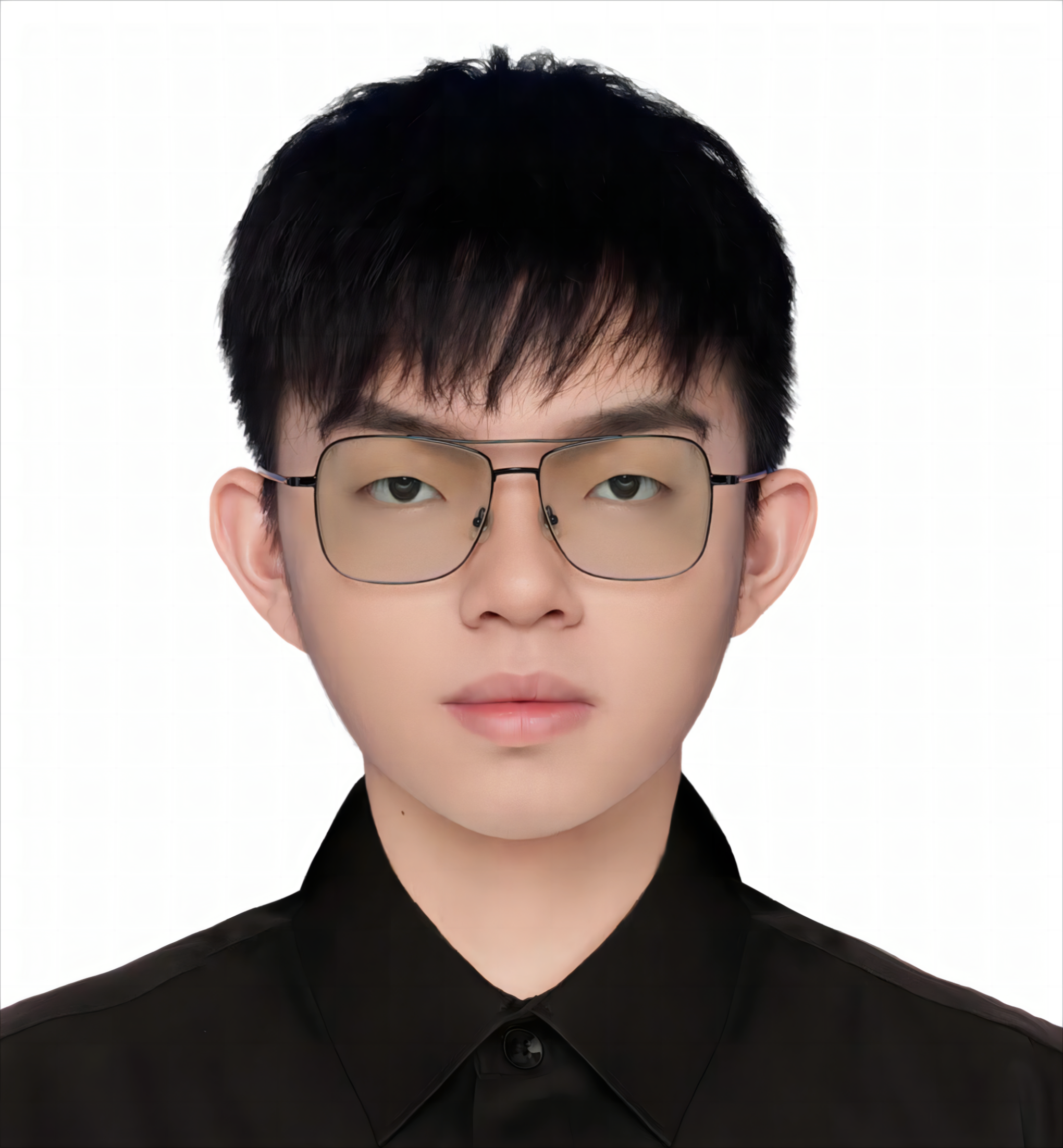}}]{Haicheng Liao}(Student Member, IEEE) received the B.S. degree in software engineering from the University of Electronic Science and Technology of China (UESTC) in 2022. He is currently pursuing the Ph.D. degree at the State Key Laboratory of Internet of Things for Smart City and the Department of Computer and Information Science, University of Macau. His research interests include connected autonomous vehicles and the application of deep reinforcement learning to autonomous driving.
\end{IEEEbiography}
\begin{IEEEbiography}
[{\includegraphics[width=1in,height=1.25in, clip,keepaspectratio]{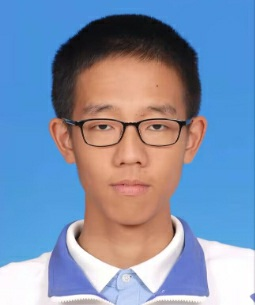}}]  {Yongkang Li}(Student Member, IEEE) is an undergraduate student in software engineering from the University of Electronic Science and Technology of China. He is currently serving as a Research Assistant at the State Key Laboratory of Internet of Things for Smart City and the Department of Computer and Information Science, University of Macau. His research interests include computer vision and trajectory predictions for autonomous driving.
\end{IEEEbiography}
\begin{IEEEbiography}
[{\includegraphics[width=1in,height=1.25in,clip,keepaspectratio]{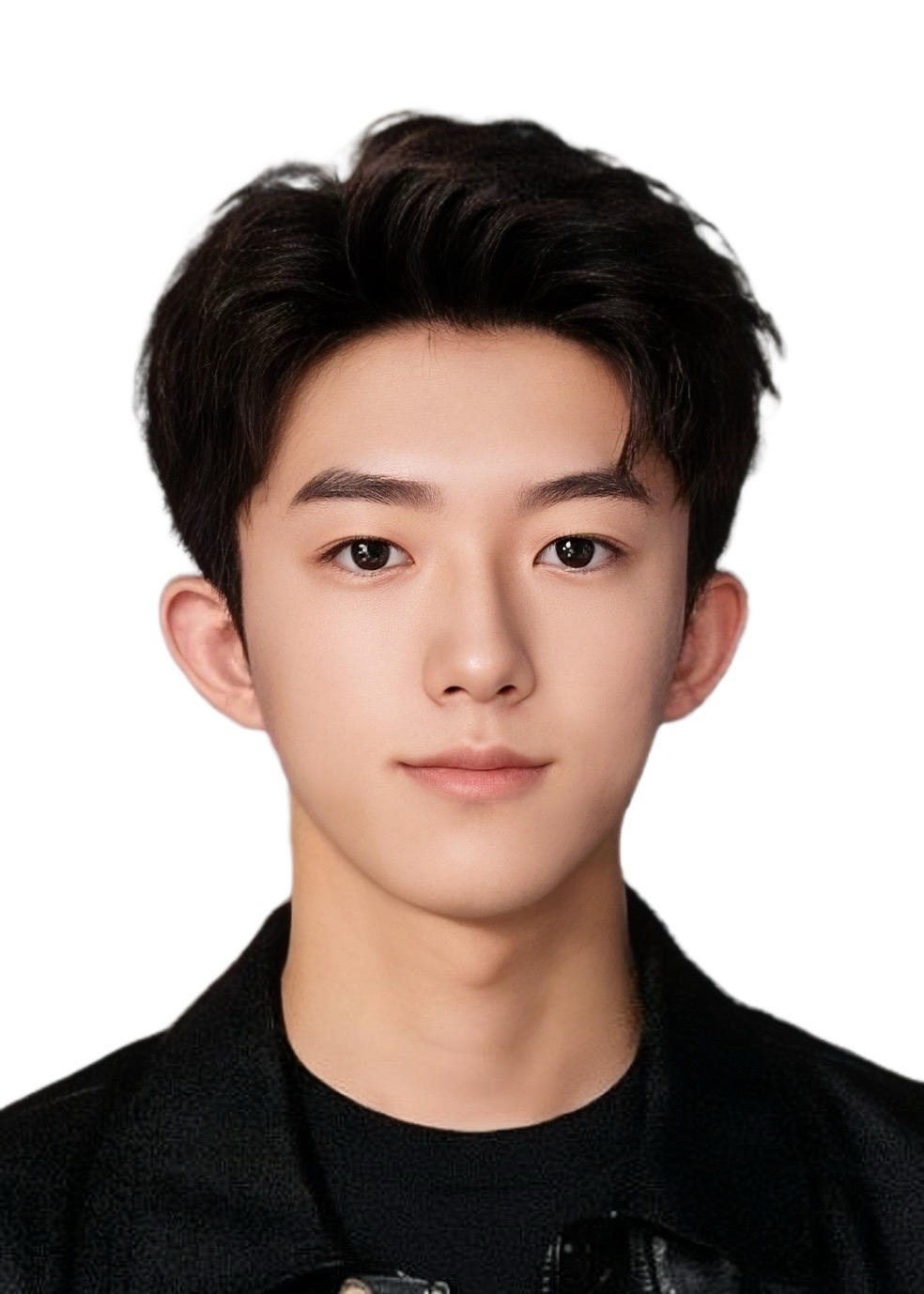}}] {Zhenning Li} (Member, IEEE) received his Ph.D. in Civil Engineering from the University of Hawaii at Manoa, Honolulu, Hawaii, USA, in 2019. Currently, he holds the position of Assistant Professor at the State Key Laboratory of Internet of Things for Smart City, as well as the Department of Computer and Information Science at the University of Macau, Macau. Over his academic career, he has published over 40 papers. His main areas of research focus on the intersection of connected autonomous vehicles and Big Data applications in urban transportation systems. He has been honored with several awards, including the TRB Best Young Researcher award and the CICTP Best Paper Award, amongst others.
\end{IEEEbiography}

\begin{IEEEbiography}
[{\includegraphics[width=1in,height=1.25in,clip,keepaspectratio]{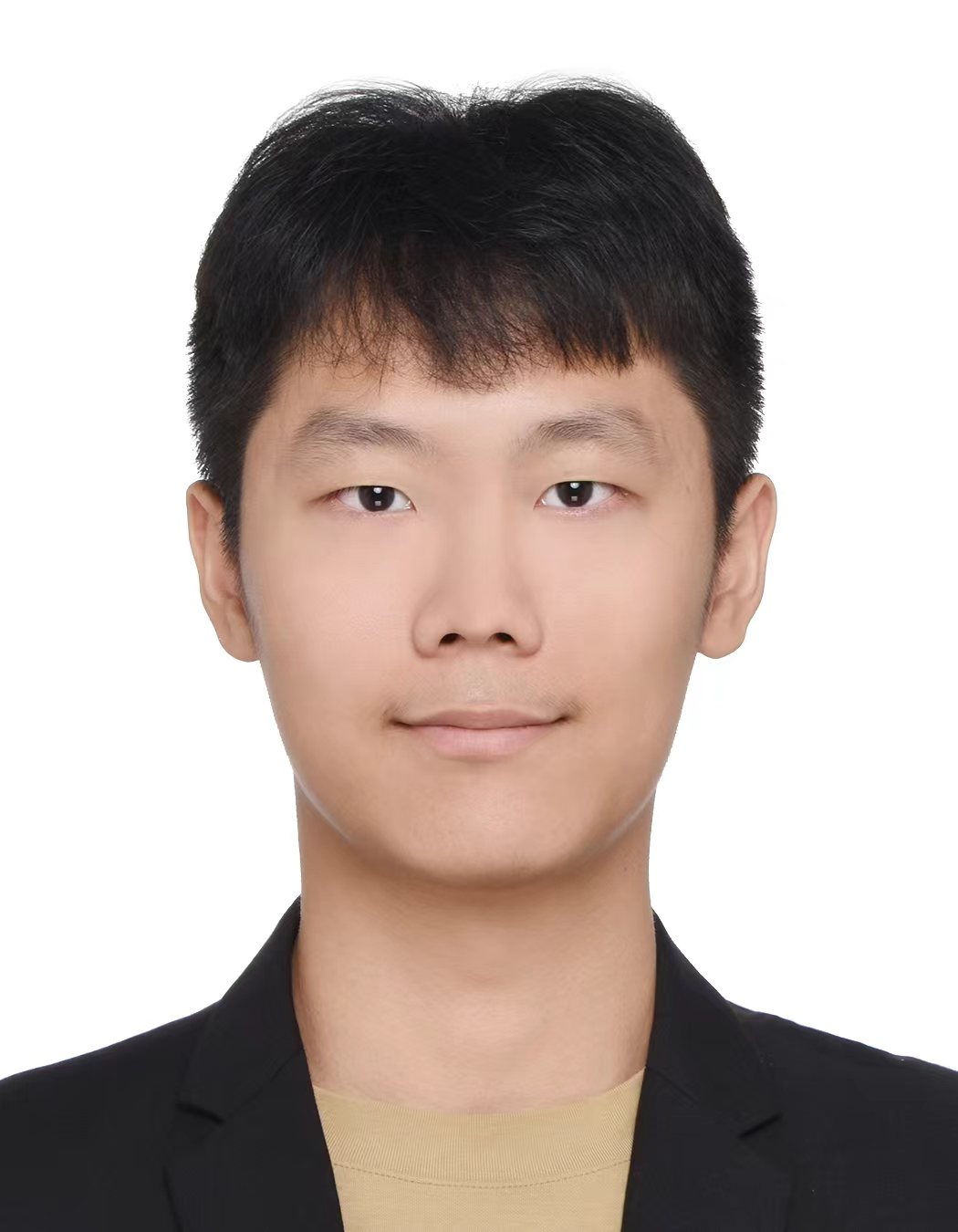}}]
{Chengyue Wang} is currently a Research Assistant at the State Key Laboratory of Internet of Things for Smart City and the Department of Civil Engineering at the University of Macau. He holds an MS degree in Civil Engineering from the University of Illinois Urbana-Champaign (2022) and a B.E in Transportation Engineering from Chang'an University (2021). His research primarily focuses on the innovative integration of artificial intelligence with autonomous driving technologies and the development of intelligent transportation systems.
\end{IEEEbiography}

\begin{IEEEbiography}
[{\includegraphics[width=1in,height=1.25in,clip,keepaspectratio]{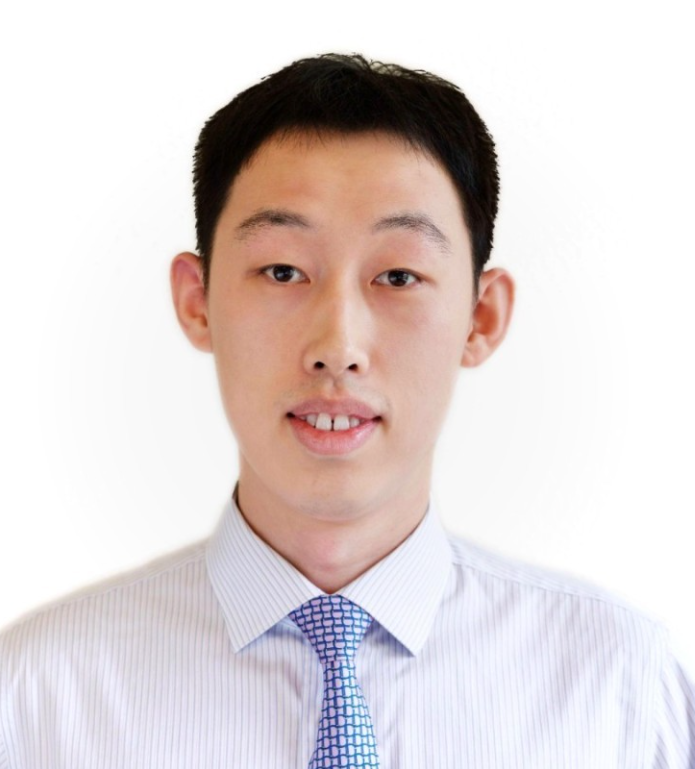}}]
{Zhiyong Cui} received the B.S. degree in software engineering from Beihang University, Beijing, China, in 2012, the M.S. degree in software engineering and microelectronics from Peking University, Beijing, in 2015, and the Ph.D. degree in civil engineering from the University of Washington, Seattle, WA, USA, in 2021. He is currently a Professor at the School of Transportation Science and Engineering, at Beihang University. He was a University of Washington Data Science Postdoctoral Fellow with the eScience Institute, Seattle, WA, USA. His primary research interests include deep learning, machine learning, urban computing, traffic forecasting, connected vehicles, and transportation data science. He was the recipient of the IEEE ITSS Best Dissertation Award in 2021 and Best Paper Award at the 2020 IEEE International Smart Cities Conference.
\end{IEEEbiography}

\begin{IEEEbiography}
[{\includegraphics[width=1in,height=1.25in,clip,keepaspectratio]{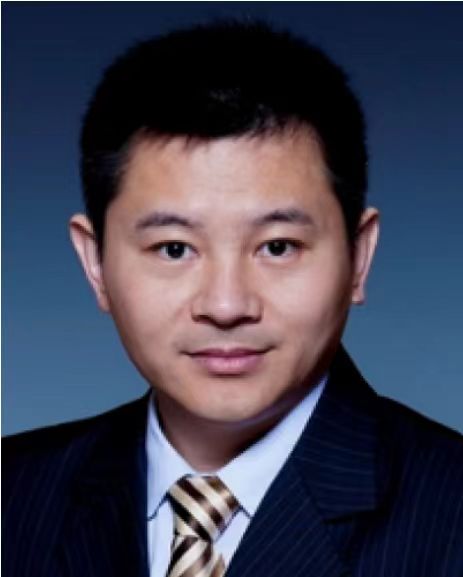}}]
{Shengbo Eben Li} (Senior Member, IEEE) received his M.S. and Ph.D. degrees from Tsinghua University in 2006 and 2009, respectively. Before joining Tsinghua University, he had worked at Stanford University, University of Michigan, and UC Berkeley. His active research interests include intelligent vehicles and driver assistance, deep reinforcement learning, optimal control and estimation, etc. He is the author of over 130 peer-reviewed journal/conference papers and the co-inventor of over 30 patents. He is the recipient of the best (student) paper awards of IEEE ITSC, ICCAS, IEEE ICUS, CCCC, etc. His important awards include the National Award for Technological Invention of China (2013), the Excellent Young Scholar of NSF China (2016), the Young Professor of ChangJiang Scholar Program (2016), the National Award for Progress in Sci \& Tech of China (2018), Distinguished Young Scholar of Beijing NSF (2018), Youth Sci \& Tech Innovation Leader from MOST (2020), etc. He also serves as the Board of Governor of the IEEE ITS Society, Senior AE of IEEE OJ ITS, and AEs of IEEE ITSM, IEEE Trans ITS, Automotive Innovation, etc.
\end{IEEEbiography}
\begin{IEEEbiography}
[{\includegraphics[width=1in,height=1.25in,clip,keepaspectratio]{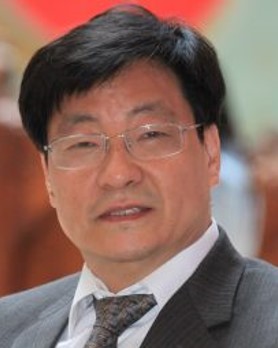}}]{Chengzhong Xu} (Fellow, IEEE) received the Ph.D. degree from The University of Hong Kong, in 1993. He is currently the chair professor of computer science and the dean with the Faculty of Science and Technology, University of Macau. Prior to this, he was with the faculty at Wayne State University, USA, and the Shenzhen Institutes of Advanced Technology, Chinese Academy of Sciences, China. He has published more than 400 papers and more than 100 patents. His research interests include cloud computing and data-driv intelligent applications. He was the Best Paper awardee or the Nominee of ICPP2005, HPCA2013, HPDC2013, Cluster2015, GPC2018, UIC2018, and AIMS2019. He also won the Best Paper award of SoCC2021. He was the Chair of the IEEE Technical Committee on Distributed Processing from 2015 to 2019.
\end{IEEEbiography}

\end{document}